\documentclass[review]{fcs}
\usepackage{amsmath}
\usepackage{amssymb}
\usepackage{mathtools}
\usepackage{amsfonts}
\usepackage{enumitem}
\usepackage{multirow}
\usepackage{makecell}
\usepackage{float}
\usepackage{graphicx}
\usepackage[shortcuts]{extdash}
\usepackage{longtable}
\usepackage[figuresright]{rotating}
\usepackage{utfsym}
\usepackage{tablefootnote}
\usepackage{footnote}
\makesavenoteenv{table}
\usepackage{colortbl}
\usepackage{threeparttable}
\title{A Comprehensive Survey of Federated Transfer Learning: Challenges, Methods and Applications}
\shorttitle{A Comprehensive Survey of Federated Transfer Learning}
\author[1]{Wei Guo}
\author*[1]{Fuzhen Zhuang}
\author*[2]{Xiao Zhang}
\author[3]{Yiqi Tong}
\author[4]{Jin Dong}
\address[1]{Institute of Artificial Intelligence, Beihang University, Beijing 100191, China}
\address[2]{School of Computer Science and Technology, Shandong University, Shandong 266237, China}
\address[3]{School of Computer Science and Engineering, Beihang University, Beijing 100191, China}
\address[4]{Beijing Academy of Blockchain and Edge Computing, Beijing 100080, China}

\fcssetup{
  received       = {month dd, yyyy},
  accepted       = {month dd, yyyy},
  corr-email     = {zhuangfuzhen@buaa.edu.cn, xiaozhang@sdu.edu.cn},
}
\begin{abstract}
\begin{sloppypar}
\noindent Federated learning (FL) is a novel distributed machine learning paradigm that enables participants to collaboratively train a centralized model with privacy preservation by eliminating the requirement of data sharing. In practice, FL often involves multiple participants and requires the third party to aggregate global information to guide the update of the target participant. Therefore, many FL methods do not work well due to the training and test data of each participant may not be sampled from the same feature space and the same underlying distribution. Meanwhile, the differences in their local devices (system heterogeneity), the continuous influx of online data (incremental data), and labeled data scarcity may further influence the performance of these methods. To solve this problem, federated transfer learning (FTL), which integrates transfer learning (TL) into FL, has attracted the attention of numerous researchers. However, since FL enables a continuous share of knowledge among participants with each communication round while not allowing local data to be accessed by other participants, FTL faces many unique challenges that are not present in TL. In this survey, we focus on categorizing and reviewing the current progress on federated transfer learning, and outlining corresponding solutions and applications. Furthermore, the common setting of FTL scenarios, available datasets, and significant related research are summarized in this survey.

\end{sloppypar}
\end{abstract}

\begin{document}
\begin{sloppypar}
\section{Introduction}
In recent years, we have witnessed breakthroughs in machine learning, especially deep neural networks (DNNs), in various fields such as computer vision, smart cities, health care, and recommendation systems, etc. Driven by high-quality training data, these methods have achieved impressive performance and even outperformed humans in certain tasks. With the rapid growth of the mobile Internet, a large amount of data is produced by billions of smart devices. However, these collected data cannot be directly uploaded to cloud servers or data centers for centralized processing due to limitations in data security, user privacy protection, and network bandwidth, which poses substantial challenges to the traditional machine learning approach. Such phenomena is commonly known as ``isolated data islands''.

One emerging paradigm for enabling distributed machine learning to solve this problem is federated learning (FL), which was first proposed by \cite{mcmahan2017communication}. The main idea of FL is to collaboratively train a centralized machine learning model with privacy preservation by transmitting and aggregating model parameters between the distributed participants, which eliminates the requirement of local data sharing and each participant can maintain ownership of their data. However, in certain FL scenarios, the data distribution varies widely between participants. For example, the training data from different participants share the same feature space but may not share the same sample ID space, or the training data from different participants may not even share the same feature space \cite{feng2022semi}. Therefore, when participants want to utilize global information to improve model utility through FL aggregation, the difference in data distributions, feature space, and label space among participants will influence the model convergence to the optimum\cite{zhang2021survey,gao2022feddc,shi2022optimization,liu2022vertical}. Furthermore, due to inconsistent local storage, computational, and communication capabilities among different participant devices, FL may grapple with system heterogeneity challenges, leading to straggler situations or high error rates. In addition to the above-mentioned data heterogeneity and system heterogeneity problems, FL also suffers from model heterogeneity, incremental data, and labeled data scarcity challenges, which are also focal points of attention among many researchers.
\begin{figure*}[t]
  \flushleft
  \includegraphics[scale=0.3]{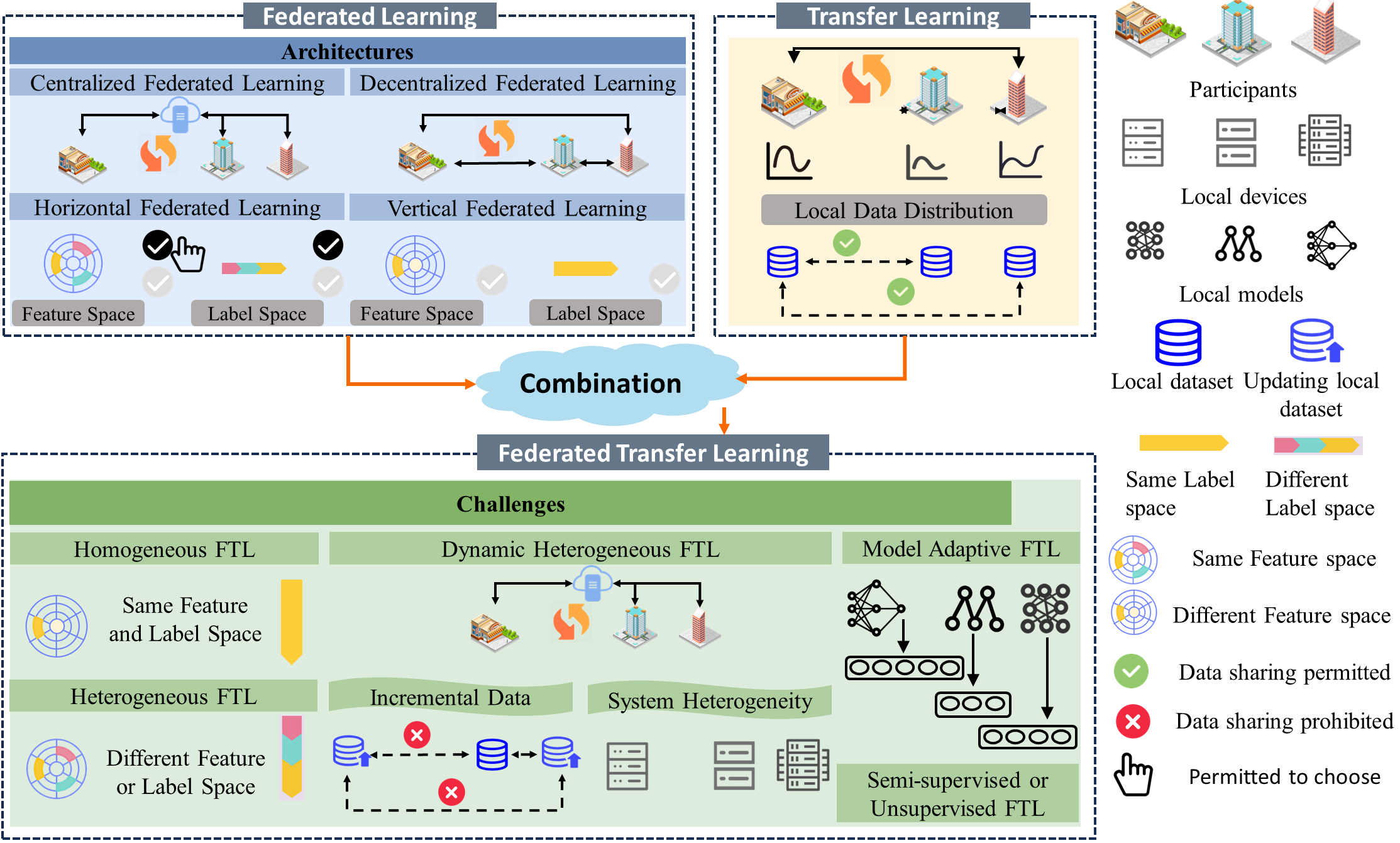}
  \caption{The overview of FTL}
  \label{fig:intro}
\end{figure*}
\begin{figure*}[t]
  \includegraphics[scale=0.28]{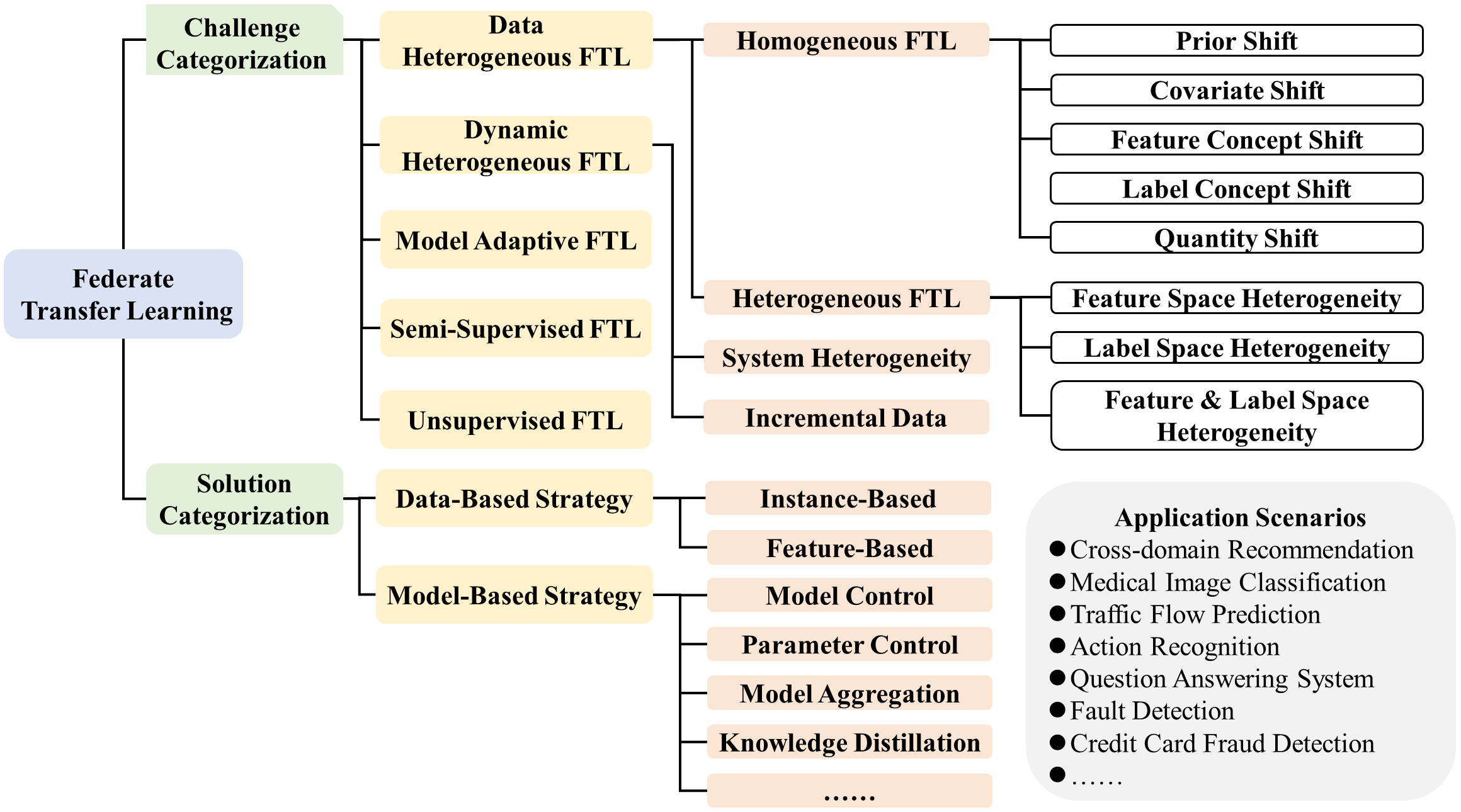}
  \caption{Categorizations of FTL.}
  \label{fig:framework}
\end{figure*}

To address the aforementioned challenges, transfer learning (TL) is employed in FL as an effective method of facilitating knowledge transfer between source and target domains \cite{zhuang2020comprehensive}. The main concept of TL is to minimize the divergence between the distributions of different domains. Similarly, in one communication round of FL, we could consider each participant as the target domain and the other participants as the source domains. Given that FL often involves multiple participants, i.e., multiple source domains, and requires the central server to aggregate information (e.g., model parameters) from multiple participants to guide the update of the target participant. In the process of continuous interaction among participants, knowledge is mutually transferred, which allows a local model obtained from a specific domain to be used by other participants through TL, thus alleviating limitations such as data heterogeneity, system heterogeneity, incremental data, and labeled data scarcity. We rethink FL in \cite{10.1145/3298981} from the perspective of TL, and refer to the combination of FL and TL as federated transfer learning (FTL) shown in Figure \ref{fig:intro}.

However, in classical TL strategies, the target domain can directly access the source domain data or model information, which contradicts the principle of FL. Hence, these TL strategies could not be directly applied in the FL. Moreover, the standard FL scheme contains a sending and receiving process through the communication between participant and server to ensure that the global model is updated and optimized across all local participants. So in a communication round, local participants can act as source and target domains at different stages. Concretely, during the sending stage, each participant acts as a source domain to transfer local knowledge to other participants. During the receiving stage, each participant serves as the target domain to receive knowledge from others. These conditions increase the difficulty of applying TL to FL.

Overall, the above unique challenges of FTL have captured the attention of numerous researchers, and many significant contributions have been made. Existing surveys in the FL field mainly focus on traditional FL \cite{rahman2021challenges,liu2022distributed,li2020survey,zhang2021survey}, including horizontal federated learning, vertical federated learning \cite{liu2022vertical}, incentive mechanism \cite{zhan2021survey}, privacy protection \cite{yin2021comprehensive,lyu2020threats}, or introducing FL applications such as healthcare \cite{pfitzner2021federated,nguyen2022federated}, mobile edge networks \cite{lim2020federated}, and internet of things (IoT) \cite{nguyen2021federated}. Despite some studies \cite{zhu2021federated,tan2022towards} focus on not identically and independently distributed (Non-IID) or other heterogeneous scenarios, such as model heterogeneity, device heterogeneity in FL, there is still a lack of systematic and comprehensive review on the definition, challenges, and corresponding solutions specific for the application of TL in FL, i.e., FTL.

To fill this gap, this survey is dedicated to giving a comprehensive survey of FTL, including definitions, a categorization, and a discussion of existing challenges and corresponding solutions, common setting scenarios of distribution heterogeneity, available datasets, as well as an outline of current FTL applications and future prospects. In detail, Figure \ref{fig:framework} shows the categorizations of FTL and corresponding solutions. Section \ref{sec:definition} demonstrates related definitions of FL and TL, we classify the common settings of FTL scenarios into six categories, including homogeneous FTL, heterogeneous FTL, dynamic heterogeneous FTL, model adaptive FTL, semi-supervised FTL, and unsupervised FTL. Section \ref{sec:data methods} - \ref{sec:usftl} systematically summarizes the corresponding solutions of existing FTL works in these scenarios, including motivation, core algorithm, model design, privacy-preserving mechanism, and communication architecture they adopt. Since some studies have involved multiple FTL scenarios, we only describe the major issues addressed by these studies. Finally, recognizing that systems and infrastructure are critical to the success of FTL, we outline current applications of FTL and propose future prospects.

The key contributions of this work are summarized as follows.
\begin{enumerate}
     \item This survey is the first to systematically and comprehensively rethink FL based on TL (FTL). We provide the definitions of FTL and its challenges including homogeneous FTL, heterogeneous FTL, dynamic heterogeneous FTL, model adaptive FTL, semi-supervised FTL, and unsupervised FTL, and further detail these challenges of FTL through examples.
     \item Based on existing FTL solutions, which include both data-based and model-based strategies, we give the current research status for FTL challenges. 
     \item We summarize the scenario settings of the homogeneous FTL shown in Table \ref{tab:settings}, which is the most common situation in FTL, including the setup methods and applied datasets. Meanwhile, to make checking convenient, we outline the existing research on FTL in Table \ref{tab:study1}, \ref{tab:study2}, \ref{tab:study3}.
\end{enumerate}

\section{Overview}
In this section, the common notations used in this survey are listed in Table \ref{tab:notation} for convenience. Besides, we further introduce the definitions, categorizations, and open challenges related to transfer learning, federated learning, and federated transfer learning.
\begin{table*}[]
\centering
\setlength{\tabcolsep}{8mm}
\renewcommand\arraystretch{0.8}
\caption{\centering The common notations.}
\label{tab:notation}
\begin{tabular}{
>{\columncolor[HTML]{C6D4E9}} c
>{\columncolor[HTML]{E8EEF6}} l
>{\columncolor[HTML]{C6D4E9}} c
>{\columncolor[HTML]{E8EEF6}} l}
\hline
\cellcolor[HTML]{A3BADC}Symbol &\multicolumn{1}{c}{\cellcolor[HTML]{A3BADC} Definition} &\cellcolor[HTML]{A3BADC} Symbol &\multicolumn{1}{c}{\cellcolor[HTML]{A3BADC} Definition} \\ \hline
$\emph{n}$ & Number of instances  & $\emph{m}$  & Number of domains\\
$\emph{k}$& Number of participants&$\bar{k}$& Actual number of participants\\
$\emph{z}$& Number of classes & $\emph{l}$  & Number of model layers\\
$\emph{p}$& Number of servers & $\emph{s}$  & Server\\
$\emph{u}$& Participant& $\emph{g}$  & Global\\
$\emph{d}$& Threshold &$\emph{f}$ & Decision function\\
$\emph{x}$& Feature vector&$\emph{y}$& Label\\
$\emph{e}$ & Communication round & $\emph{F}$ & Device \\
$\emph{A}$& Active participant & $\emph{B},\emph{C}$&Passive participant\\
$\emph{D}$ & Domain & $\mathcal{T}$ & Task \\
$\mathcal{X}$& Feature space&$\mathcal{Y}$& Label space\\
$\emph{X}$ & Instance space & $\emph{I}$& Sample ID space\\
$\emph{Y}$ & Label set corresponding to X & $\emph{S}$& Source domain\\
$\emph{T}$ & Target domain & $\emph{L}$  & Labeled instances\\
$\emph{U}$ & Unlabeled instances & $\mathcal{L}$& Loss function \\
$\mathcal{R}$ & Relationship matrix &$\mathcal{M}$& Model\\
$\mathcal{E}$ & Extractor &$\mu$ & Mean \\
$\sigma$ & Variance&$\lambda$ & Importance variable\\
$\delta$ & Tradeoff parameter & $\rho$ & Interpolation coefficient\\
$\Omega$ & Structural risk &$\theta$  & Model parameters \\ \hline  
\end{tabular}
\end{table*}

\subsection{Definition}
\label{sec:definition}
Following with previous works \cite{zhuang2020comprehensive,10.1145/3298981}, we first give the definitions of 
``domain'', ``task'', ``transfer learning'', and ``federated learning'' that are used in this survey, respectively. The involved common notations are summarized in the Table \ref{tab:notation}. 

\noindent $\textbf{Definition 1.}$ (Domain) A domain $\emph{D}$ is constituted by two elements: a feature space $\mathcal{X}$ and an probability distribution $\emph{P(X)}$, where the symbol $\emph{X}$ represents an instance set, i.e., $\emph{X} = \{x|x_{i}\in \mathcal{X}, \emph{i} = 1,...,n\}$. Thus, a domain $\emph{D}$ can be denoted as $\emph{D} = \left\{\mathcal{X},P(X)\right\}$.  In general, if two domains are different, then they may have different feature spaces $\mathcal{X}$ or different probability distributions $\emph{P(X)}$ \cite{pan2009survey}.

\noindent $\textbf{Definition 2.}$ (Task) A task $\mathcal{T}$ is constituted by a label space $\mathcal{Y}$ and a decision function $f$, denoted as $\mathcal{T} = \left\{\mathcal{Y}, f\right\}$. Given the training data, the decision function $f$ is used to predict the corresponding label $y \in \mathcal{Y}$, where $f$ is not explicit but can be inferred from the sample data.

\noindent $\textbf{Definition 3.}$ (Transfer Learning) Given an/some observation(s) corresponding to $\emph{m}^{S} \in \mathbb{N}^{+}$ source domain(s) and task(s) (i.e., $\{(\emph{D}_{S_{i}},\mathcal{T}_{S_{i}})|i=1,...,m^{S}\})$), and an/some observation(s) about $\emph{m}^{T} \in \mathbb{N}^{+}$ target domain(s) and task(s) (i.e., $\{(\emph{D}_{\mathcal{T}_{j}},\mathcal{T}_{T_{j}})|j=1,...,m^{T}\}$), transfer learning aims to utilize the knowledge implied in the source domain(s) to improve the performance of the learned decision functions $f^{T_{j}}(j=1,...,m^{T})$ on the target domain(s) \cite{zhuang2020comprehensive}.

\noindent $\textbf{Definition 4.}$ (Federated Learning) Assume there are $\emph{k}$ participants {$\emph{u}_{1},...,\emph{u}_{k}$}, each aiming to train a machine learning model with their own private datasets {$\emph{X}_{1},..$ $.,\emph{X}_{\emph{k}}$}. A conventional approach is to upload all data together and use $\emph{X}=\emph{X}_{1} \cup ... \cup \emph{X}_{k}$ to train a global model $\mathcal{M}_{SUM}$. However, in many application scenarios, participants cannot directly upload their own data or access the data of other participants. Therefore, a typically federated learning system is a distributed learning process in which the participant collaboratively train a model $\mathcal{M}_{FED}$ without sharing local private data $\emph{X}_{i}$ \cite{10.1145/3298981}.

\noindent $\textbf{Definition 5.}$ (Federated Transfer Learning) Given there are some challenges in FL for participants $i$ and $j$ ($i = 1, ..., k$), including data heterogeneity ($\mathcal{X}_{i} \neq \mathcal{X}_{j}$ or $\mathcal{Y}_{i} \neq \mathcal{Y}_{j}$ or $P_{i}(X_{i},Y_{i}) \neq P_{j}(X_{i},Y_{j})$, where instance space $X$ consists of feature space $\mathcal{X}$ and label space $\mathcal{Y}$), system heterogeneity ($F_{i} \neq F_{j}$), incremental data ($X^{i}_{e-1} \neq X^{i}_{e}$), and scarcity of labeled data ($\bigcup_{i}X_{i} = X^{L} \rightarrow 0$), the FL combines the TL to solve these challenges, called FTL. The specific definition of FTL as follows. Given $k$ participants $u_{1},...,u_{k}$ in FL, the central server set $\{s_{p} | p = 1, ..., \mathcal{N}^{+}\}$ is designed to achieve model convergence over the $E$ communication rounds. 

During each communication round, the typical federated transfer learning process includes two distinct stages:
\begin{enumerate}
    \item Sending stage: participant(s) $u_{i}$ ($1 < i \leq k$) is(are) assumed as the role of the source domain(s) $D^{S_{i}}$ ($1 < i \leq m^{S}$), where they are responsible for contributing local an/some observation(s) ($D_{S_{i}}$, $\mathcal{T}_{S_{i}}$) corresponding to $D_{S_{i}}$ and task(s) $\mathcal{T}_{S_{i}}$ to the central server $s_{p}$ without sharing local raw data {$\emph{X}_{1},..$ $.,\emph{X}_{\emph{k}}$}. The server then leverages the collected sending information to implement the aggregation process.
    \item Receiving stage: once the aggregation is complete, participant(s) $u_{j}$ ($1 < j \leq k$) then are assumed as the role of the target domain(s) $\{(\emph{D}_{\mathcal{T}_{j}},\mathcal{T}_{T_{j}})|j=1,...,m^{T}\}$ and utilize the received global aggregation information to perform local model updates.
\end{enumerate}
The above sending and receiving stages are assumed to repeat for $E$ communication rounds, or until the model is observed to converge. Particularly, when $p = 0$, the above process is considered as a decentralized federated transfer learning process.

\subsection{Category of federated learning}
According to the characteristics of data distribution among connected participants, FL can be categorized into horizontal FL (HFL) and vertical FL (VFL). Generally, HFL considers the distributed participants to have data with the same features but are different in sample space, while VFL considers the distributed participants to have the same samples but different features to jointly train a global model \cite{zhu2021federated,liu2022vertical}. Federated transfer learning in \cite{10.1145/3298981} refers that these participants have differences in both feature space and label space. Due to the limited research on federated transfer learning in \cite{10.1145/3298981}, this survey categorizes federated transfer learning and VFL as a type of VFL for description.

On the other hand, depending on whether there is a/some central server(s) responsible for coordinating participants, FL can also be divided into centralized FL (CFL) and decentralized FL (DFL), where CFL assumes that there is a/some server(s) to gather local model-related information or other training information from the participants and then distributes the updated global model back to the participants, while DFL assumes participants directly aggregate information from neighboring participants \cite{10.1145/3298981}. In the following, we will provide a brief introduction to these FL frameworks and discuss the various settings of source domains, target domains, and tasks when employing transfer learning within these frameworks.

\begin{figure*}[t]
  \centering
  \includegraphics[scale=0.23]{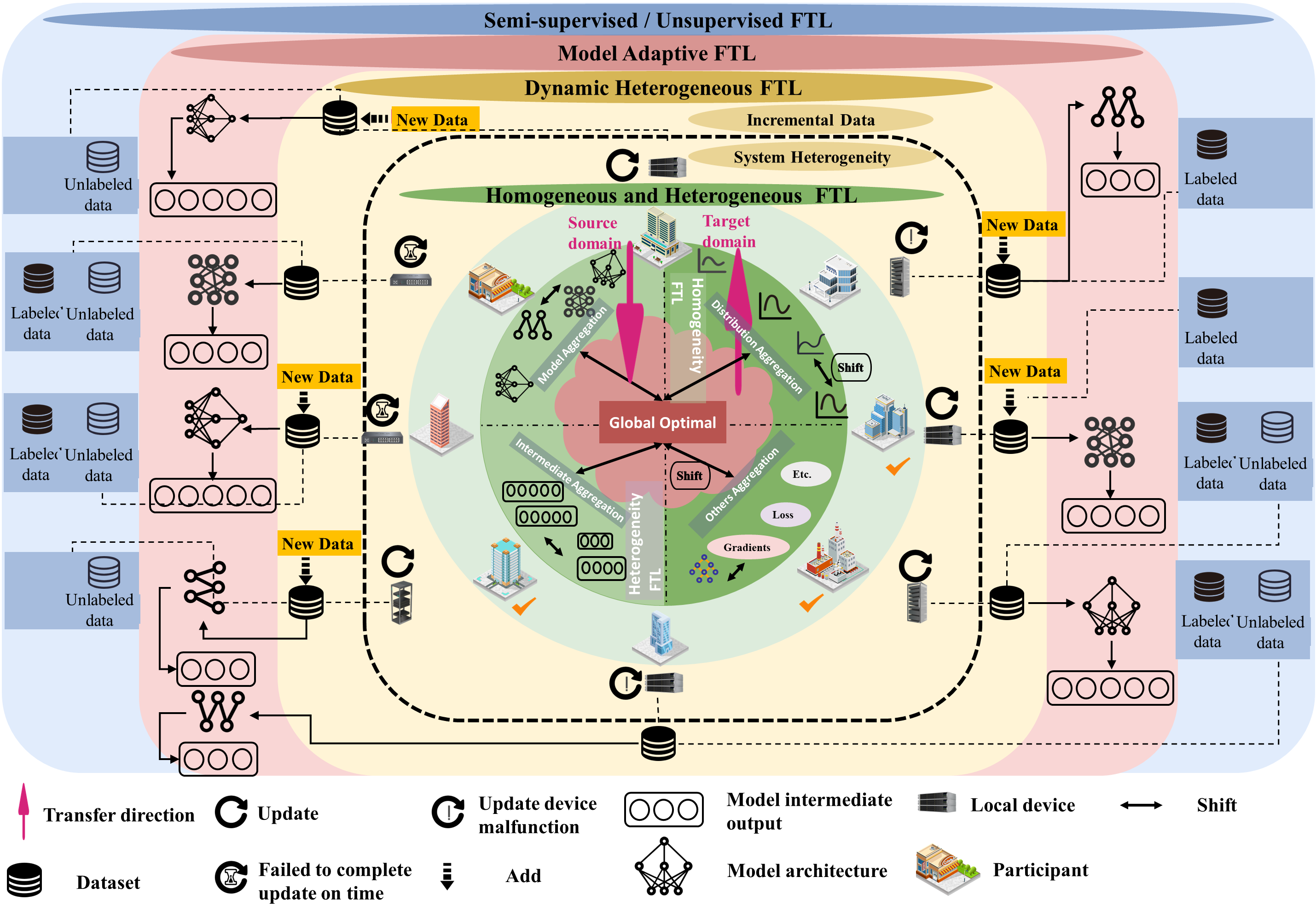}
  \caption{The challenges of FTL}
  \label{fig:FTL}
\end{figure*}
\subsubsection{Horizontal federated learning}
HFL is commonly found in scenarios where participants share the same feature space $\mathcal{X}$ but different sample space $\emph{I}$, which meets homogeneity FTL described in Section \ref{sec:homo FTL}. For example, the medical record data of two regional hospitals $i$ and $j$ may be very similar due to they use the same information system, which both record the patient's name, age, gender, and other user private data, so their feature spaces are the same ($\mathcal{X}_{i} = \mathcal{X}_{j}$). However, the two hospitals have different user groups ($\emph{I}_{i} = \emph{I}_{j}$) from their respective regions, and the user intersection of their local datasets is very limited. In FTL, any participant can serve as a source domain ($\emph{S}$) to provide knowledge or as a target domain ($\emph{T}$) to receive knowledge from other participants in the same feature space, therefore, we define HFL in homogeneous FTL as:
\begin{equation*}
    \mathcal{X}^{S}_{i} = \mathcal{X}^{T}_{j}, \mathcal{Y}^{S}_{i} = \mathcal{Y}^{T}_{j}, \emph{I}^{S}_{i} \neq \emph{I}^{T}_{j}, \forall \emph{X}^{S}_{i},\emph{X}^{T}_{j}
\end{equation*}
\begin{equation*}
    or\, \mathcal{X}^{T}_{i} = \mathcal{X}^{S}_{j}, \mathcal{Y}^{T}_{i} = \mathcal{Y}^{S}_{j}, \emph{I}^{T}_{i} \neq \emph{I}^{S}_{j}, \forall \emph{X}^{T}_{i},\emph{X}^{S}_{j}, i \neq j.
\end{equation*}
From an extended perspective, HFL meets heterogeneous FTL when participants' label space is inconsistent in the knowledge-transferring process, the HFL in heterogeneous FTL can be represented as:
\begin{equation*}
    \mathcal{X}^{S}_{i} = \mathcal{X}^{T}_{j}, \mathcal{Y}^{S}_{i} \neq \mathcal{Y}^{T}_{j}, \emph{I}^{S}_{i} \neq \emph{I}^{T}_{j}, \forall \emph{X}^{S}_{i},\emph{X}^{T}_{j}
\end{equation*}
\begin{equation*}
    or\, \mathcal{X}^{T}_{i} = \mathcal{X}^{S}_{j}, \mathcal{Y}^{T}_{i} \neq \mathcal{Y}^{S}_{j}, \emph{I}^{T}_{i} \neq \emph{I}^{S}_{j}, \forall \emph{X}^{T}_{i},\emph{X}^{S}_{j}, i \neq j.
\end{equation*}

\subsubsection{Vertical federated learning}
Unlike HFL where all participants have their own local data labels, in the VFL scenario, participants' feature spaces $\mathcal{X}$ are inconsistent, and their sample spaces $\emph{I}$ may also not be entirely the same. For example, suppose there is a high degree of overlap in the customer groups $\emph{I}$ between a bank $i$ and a telecommunications company $j$ in the same region. The bank $i$ has information on users' credit history ($\mathcal{X}_{i}$), such as loan repayment details and credit card usage, while the telecommunications company holds data on users' call logs, data usage, and payment records ($\mathcal{X}_{j}$), where the feature space is different ($\mathcal{X}_{i} \neq \mathcal{X}_{j}$). These two entities, which all act both as source domains or as target domains, can engage in VFL to mutually enhance their services in different feature spaces. We summarize VFL in heterogeneous FTL as: 
\begin{small}
\begin{equation*}
    \mathcal{X}^{S}_{i} \neq \mathcal{X}^{T}_{j}, \mathcal{Y}^{S}_{i} = (\neq) \mathcal{Y}^{T}_{j}, \emph{I}^{S}_{i} = (\neq) \emph{I}^{T}_{j}, \forall \emph{X}^{S}_{i},\emph{X}^{T}_{j}
\end{equation*}
\end{small}
\begin{small}
\begin{equation*}
    or\, \mathcal{X}^{T}_{i} \neq \mathcal{X}^{S}_{j}, \mathcal{Y}^{T}_{i} = (\neq) \mathcal{Y}^{S}_{j}, \emph{I}^{T}_{i} = (\neq) \emph{I}^{S}_{j}, \forall \emph{X}^{T}_{i},\emph{X}^{S}_{j}, i \neq j.
\end{equation*}
\end{small}

\subsubsection{Centralized federated learning}
Standard CFL requires one or more central servers to build a global model by collecting local information from distributed participants \cite{konevcny2016federated}, which involves three fundamental steps as described below:

\begin{enumerate}
\item Receiving stage: each participant receives the initial model sent by the server.
\item Sending stage: participants $\emph{u}_{i}$ use their own private data $\emph{X}_{i}$ to train the local model (add local model notion), and then send the local model to the server.
\item Receive stage: The central server updates the global model $\mathcal{M}_{g}$ by collecting and aggregating all the local updates and then sends the updated global model back to the participants.
\end{enumerate}
In the FTL setting, during the sending and receiving stages, participants share knowledge through a central aggregation strategy, where each participant can act as a source domain providing knowledge or a target domain receiving knowledge. For instance, during the sending stage, participants act as source domains providing model parameters, while the server acts as the target domain, aggregating these parameters to form a global model. Conversely, in the receiving stage, the server serves as the source domain providing global model parameters to each participant.

\subsubsection{Decentralized federated learning}
Compared with CFL, DFL is conducted over different participants $\left\{u_{1},...,u_{\emph{k}}\right\}$ without a central parameter server for global model aggregation. Each participant uses a private local dataset to optimize their local model after receiving model updates from other participants. This process involves two fundamental steps as described below:

\begin{enumerate}
    \item Receiving stage: participant $\emph{u}_{i}$ federally train its initial model $\mathcal{M}_{i}$ locally with its own dataset $\emph{X}_{i}$, and then send the model $\mathcal{M}_{i}$ to other participants $\left\{u_{1},...,u_{\emph{k-1}}\right\}$ without direct data exposure.
    \item Sending stage: participant $\emph{u}_{i}$ obtain the aggregated model $\mathcal{M}_{g}$ by aggregating the received local model $\{\mathcal{M}_{1},...,$ $\mathcal{M}_{k}\}$, and then update local model with aggregated model.
\end{enumerate}
Similar to CFL, each participant in DFL could still serve as either a source domain or a target domain without a central server during different stages of the FL process.

\subsection{Federated transfer learning}
\label{sec: FTL challenges}
Transfer learning has achieved remarkable success by enabling the application of knowledge from one domain to improve performance in another, significantly reducing the need for extensive data collection and training time in new tasks \cite{long2014domain,long2015learning,bengio2012deep,raina2007self}. However, as shown in Figure \ref{fig:FTL}, constrained by the unique distributed learning paradigm of FL, current FTL studies face many additional challenging situations, including homogeneous FTL, heterogeneous FTL, dynamic heterogeneity FTL, model adaptive FTL, semi-supervised FTL and unsupervised FTL. More descriptions are presented below.

\subsubsection{Homogeneous federated transfer learning}
\label{sec:homo FTL}
Assume that the local data of participant $i$ and participant $j$ constitute a source domain $\emph{D}_{i}$ and a target domain $\emph{D}_{j}$, respectively. $\emph{D}_{i} \neq \emph{D}_{j}$ represents a difference in either their feature spaces ($\mathcal{X}_{i} \neq \mathcal{X}_{j}$) or marginal distributions ($\emph{P}_{i}(X)$ or $\emph{P}_{j}(Y)$). Similarly, if the task between participant $i$ and participant $j$ is not same, that is $\mathcal{T}_{i} \neq \mathcal{T}_{j}$, then there is a difference in their label spaces ($\mathcal{Y}_{i} \neq \mathcal{Y}_{j}$) or in their conditional distribution ($\emph{P}_{i}(y|x) \neq \emph{P}_{j}(y|x)$). 

HOFTL refers to differences in marginal distributions ($\emph{P}_{i}(X)$ or $\emph{P}_{j}(Y)$), conditional distributions ($\emph{P}_{i}(y|x) \neq \emph{P}_{j}(y|x)$), or sample sizes $n_{i} \neq n_{j}$ between participant data, which is often caused by diversity in domain or task between participants. Based on this, HOFTL includes five scenarios:
\paragraph{Prior shift: $P_{i}(\emph{Y}) \neq P_{j}(\emph{Y})$}
\paragraph{Covariate shift: $P_{i}(\emph{X}) \neq P_{j}(\emph{X})$}
\paragraph{Feature concept shift: $P_{i}(\emph{x}|\emph{y}) \neq P_{j}(\emph{x}|\emph{y})$}
\paragraph{Label concept shift: $P_{i}(\emph{y}|\emph{x}) \neq P_{j}(\emph{y}|\emph{x})$}
\paragraph{Quantity shift: $n_{i} \neq n_{j}$}
Specifically, as described in subsection 2.2.1, horizontal federated learning assumes that participants have the same feature space, so HOFTL is a form of transfer learning under this HFL assumption. Moreover, HOFTL can also be presented in vertical federated learning when there is partial overlap in the feature space between participants. Unless specifically stated, the HOFTL methods discussed in this survey are all related to HFL. Overall, compared with homogeneous transfer in traditional transfer learning that only considers marginal and conditional probability distributions, HOFTL also considers changes in the total sample size, which corresponds to the Non-IID data setting in federated learning. The detailed challenges of HOFTL are described below, and considering that homogeneous FTL is one of the most frequently discussed, we have summarized the specific settings for each homogenous FTL scenario demonstrated in Table \ref{tab:settings}.

\paragraph{Prior shift}
Prior shift, also known as class imbalance, implies that the prior probability distribution $P(Y)$ could be inconsistent between different participants when the conditional probability distribution $\emph{P}(y|x)$ is consistent \cite{kairouz2021advances}. In FL, the prior probability distribution inconsistency may occur when different participants have different class distributions in their local datasets. If these differences are not properly handled, they can lead to a federated model that performs unfair and suboptimal performance. For example, an FL system is designed to improve predictions for a specific disease (e.g., diabetes) across different hospitals that participate in model training without sharing private patient records. Hospital A is located in an urban area with a high prevalence of diabetes, possibly due to lifestyle factors prevalent in the population it serves. As a result, in hospital A's patient data, 30\% of patients might have diabetes. On the other hand, hospital B serves a rural area with a different demographic and lifestyle, resulting in only 10\% of its patients having diabetes. This difference in the prevalence of diabetes is a classic example of prior probability shift. In some extreme cases, hospital B may even have no positive cases for this disease.

\paragraph{Covariate shift}
Covariate shift, or feature distribution imbalance, describes a situation where the input feature distribution $P(X)$ is varied between participants while the conditional probability $P(y|x)$ remains consistent. This presents a unique challenge in FL because the global model is trained on data from multiple participants, and each participant's local data may represent a different underlying distribution of input features. For example, the patient population at hospital A had a higher average body mass index (BMI), which is a known risk factor for diabetes, while the patient population at hospital B had a lower average BMI. There is a significant difference in the input data (in this case, the BMI distribution) between the two hospitals, known as covariate shift.


\paragraph{Feature concept shift}
Concept drift, which includes feature concept shift and label concept shift, refers to the change in the relationship between variables $x$ and $y$, where feature concept shift implies to $\emph{P}(x|y)$ discrepancy among participants with the same prior distribution $\emph{P}(y)$ \cite{kairouz2021advances}. This type of shift can be particularly challenging in federated learning because models need to generalize across all participants' data. For example, consider two hospitals A and B jointly predicting the incidence of diabetes, where "$x$" represents the patient's health characteristics and "$y$" represents the presence or absence of diabetes. Hospital A's diabetic population mostly has a higher socioeconomic status, resulting in a different set of health characteristics, such as better control of blood sugar levels and fewer complications. In contrast, patients with diabetes at hospital B may have lower socioeconomic status and poorer health characteristics, such as uncontrolled blood sugar levels and higher rates of complications. Differences in the distribution of health characteristics ($x$) for a given diabetic patient ($y$) are an example of a $\emph{P}(x|y)$ shift.


\paragraph{Label concept shift}
Similar to feature concept shift, the label concept drift refers to $\emph{P}(y|x)$ inconsistent among participants with the same covariate distribution $\emph{P}(x)$ \cite{kairouz2021advances}. Some external events or changes may lead to changes of $\emph{P}(y|x)$ in either the source/target domain, which further renders the models from the source/target domain no longer suitable for tasks in the target/source domain. For example, in a federated recommendation system, geographical location is commonly used as the input feature to predict users' favorite items. Thus, if the emergence of tendentious policy or new pillar industries supports the economic development of area $A$, the consumption level of $A$ will be improved. In this situation, the expected user preference will change, causing the prediction results of participant $u_{i}$ from $A$ to become invalid and unsuitable for the improvement of model predictions from other regions' participants.

\renewcommand{\dblfloatpagefraction}{.7}
\begin{table*}[]
\renewcommand{\thempfootnote}{\arabic{mpfootnote}}
\centering
\setlength{\tabcolsep}{0.01mm}
\renewcommand\arraystretch{0.9}
\caption{\centering Data heterogeneity settings of HOFTL and HEFTL}
\begin{tabular}{ccccl}
\hline
\multicolumn{2}{c}{Problem Categorization}                           & Setting  &Reference  &Dataset        \\ \hline
\multirow{12}{*}{HOFTL}   & \multirow{7}{*}{Prior shift}         &Fixed ratio &\cite{younis2022fly,wu2020fedhome,jeong2018communication,li2021sample,chen2023elastic,chen2020fedbe,seo202216} &\multirow{19}{*}{\makecell[l]{CIFAR-10\footnotemark[1], CIFAR-100\footnotemark[1], \\MNIST\footnotemark[2], Tiny-Imagenet\footnotemark[3], \\ImageNet\footnotemark[3], FEMNIST\footnotemark[4], \\ OFFICE\cite{gong2012geodesic}, DIGIT\cite{peng2019moment}, \\OpenImage\cite{kuznetsova2020open}, WESAD\cite{cassara2022federated}, \\ KDD99\footnotemark[5], SVHN\footnotemark[6], HAR\footnotemark[7],\\OFFICE-Caltech 10\footnotemark[8],  \\MIMIC-III\footnotemark[9], Shakespeare\cite{mcmahan2017communication}, \\DomainNet\footnotemark[10], NSL-KDD99\footnotemark[11], \\CINIC10\cite{darlow2018cinic}, CelebA\cite{liu2015deep}, \\ StackOverflow \cite{reddi2020adaptive}}}\\
& &Natural partition &\cite{liu2021fedirm,sattler2020clustered,qayyum2022collaborative,mansour2020three,huang2019patient,ouyang2022clusterfl,duan2020self,zhang2023federated,hu2023federated,jeong2020federated,li2022learning,qi2023fedsampling,chen2020joint,deng2021auction,yang2020age,su2022online,wu2022communication}&\\
& &1 class/participant &\cite{Tuor2020DataSF,duan2020fedgroup,cassara2022federated,nagalapatti2021game}&\\
& &\textgreater 1 classes/participant &\cite{yoon2021fedmix,hao2021towards,liang2020think,zhang2023federated,briggs2020federated,duan2020fedgroup,liu2021pfa,tan2022fedproto,shen2020federated,arivazhagan2019federated,fallah2020personalized,li2020federated,deng2020adaptive,t2020personalized,hanzely2020federated,dinh2020federated,hanzely2020lower,li2021ditto,karimireddy2020scaffold,ma2021fast,collins2021exploiting,oh2021fedbabu,jang2022fedclassavg,zhuang2021collaborative,qu2023prevent,wang2023flexifed,li2021fedmask,yang2022personalized,diao2020heterofl,mcmahan2017communication,zeng2021fedcav,zhu2023confidence,lin2022federated,beilharz2021implicit,zhang2020personalized,liu2023feddwa,wang2020optimizing,nishio2019client,luping2019cmfl,xia2020multi,yang2021federatedreduction,chai2020tifl,li2021fedsae,cox2022aergia,dong2022federated,makhija2022architecture,huang2021personalized,zhu2022fednkd,chai2021fedat,hu2022spread,zhang2022personalized,nguyen2022feddrl,yoon2021federated,qu2022context,marfoq2022personalized,huang2022few,itahara2021distillation,zhang2023towards,zhuang2022divergence,wang2023fedftha}&\\
& &Dirichlet Distribution &\makecell[c]{\cite{gong2022preserving,lin2020ensemble,lin2021semifed,lubana2022orchestra,liang2022rscfed,li2023class,zhang2022fedzkt,zhou2023fedfa,yue2022neural,long2023multi,lubana2022orchestra,li2021model,chen2021bridging,yao2020continual,wang2020tackling,yu2022spatl,liu2022completely}, \\ \cite{jang2022fedclassavg,ilhan2023scalefl,liu2022deep,liu2023feddwa,cho2020client,huang2020efficiency,marfoq2022personalized,wang2023dafkd,gong2022federated,chen2020fedbe,li2020practical,sattler2020communication,zhu2021data,zhang2022fine,yang2023fedack,li2022fedhisyn,han2022fedx}}&\\ 
&&JensenShannon divergence&\cite{li2022pyramidfl}&\\
& &Half-normal distribution  &\cite{duan2020self,zhang2021dubhe}&\\ 
&& Log-normal
distribution&\cite{collins2021exploiting}&\\ \cline{2-4}
                                   & \multirow{2}{*}{Covariate shift}                &1 domain/participant   & {\begin{tabular}[c]{@{}l@{}}\cite{liu2021fedirm,zhou2023fedfa,chen2023fraug,liu2021pfa,liu2021feddg,li2021fedbn}, \\ \cite{yang2021federated,wang2022fedkc,wang2019federated,li2020federated,li2021ditto,pillutla2022federated,li2021hermes,chen2021bridging,mcmahan2017communication},  \\ \cite{zhuang2021joint,zhang2023federateddomain,zhu2023confidence,beilharz2021implicit,luping2019cmfl,ruan2022fedsoft,xie2023federated,donahue2021optimality,dai2020federated,marfoq2022personalized,li2019fedmd,li2021fedh2l,wu2021fedcg,niu2023mckd}\end{tabular}}
                                   &\\
                                   & &Mixed domain/participant &\cite{huang2023rethinking,wang2020federated}& \\ \cline{2-4}
                                   & Feature concept shift           & 1 degree/participant  &\cite{ghosh2020efficient} & \\ \cline{2-4} 
                                   & Label concept shift & &\cite{zhu2023confidence,li2019fedmd} & \\ \cline{2-4} 
                                   & \multirow{3}{*}{Quantity shift} &Natural &
                                   \makecell[c]{\cite{zhu2023confidence,qi2023fedsampling,deng2021auction,yang2021federatedreduction,tu2021feddl,ruan2022fedsoft,donahue2021model}, \\ \cite{donahue2021optimality,nguyen2022feddrl,itahara2021distillation,gao2019privacy}}
                                   & \\  
                                   & & By data source &\cite{zhou2023fedfa,yoon2021fedmix,mcmahan2017communication,zhuang2021joint}  &     \\ 
                                   &&By parameter &\cite{zeng2021fedcav}&\\
                                   \cline{1-5}
\multirow{4}{*}{HEFTL} & \multirow{2}{*}{Feature space hetergeneity}     & Overlapped feature &\cite{kang2022privacy,fu2023feast,banerjee2021fed,cassara2022federated,liu2022completely,diao2020heterofl}&\multirow{4}{*}{\makecell[l]{CIFAR-10\footnotemark[1], CIFAR-100\footnotemark[1], \\MNIST\footnotemark[2], MovieLens \cite{wu2022practical}, \\ModelNet\cite{he2022hybrid}, FEMNIST\footnotemark[4]\\NUS-WIDE\cite{wu2022practical},}}\\
& &Non-overlapped feature &\cite{feng2020multi,feng2022vertical,jiang2022vf,castiglia2023less,wu2022practical}& \\ \cline{2-4} 
                                   & Label space heterogeneity       &     &        &         \\ \cline{2-4} 
                                   & {\begin{tabular}[c]{@{}l@{}}Feature and label \\ space heterogeneity\end{tabular}} &    &   &                \\ \hline
\end{tabular}
\label{tab:settings}
\end{table*}

\paragraph{Quantity shift}
Different from the prior shift, quantity shift refers to the situation where there is a significant imbalance in the number of training samples available among participants. In FL, some participants might have a large dataset, while others may have a relatively small one. This can lead to a situation where the global model is disproportionately influenced by participants with more data, potentially leading to biases or overfitting to the characteristics of those datasets. For example, a large-scale hospital may have thousands of patient records, while a small clinic may only have a few hundred. This difference in data volume is a classic example of quantity shift in FL.

In summary, with uniform feature and label spaces, participants in homogeneous FTL still face data distribution shift problems, including prior shift, covariate shift, feature concept shift, label concept shift, and quantity shift. Most current FTL studies focus on prior and quantity shifts, with few studies tackling covariate shifts. Feature concept shift and label concept shift are even less explored. However, external elements change, like time or policy, may change the relationship between features and labels for partial participants while leaving others unchanged. This intensifies the feature concept drift and label concept drift among participants, which is worth deeper study in the future.
\footnotetext[1]{https://www.cs.toronto.edu/$\sim$kriz/cifar.html}
\footnotetext[2]{https://www.kaggle.com/datasets/hojjatk/mnist-dataset}
\footnotetext[3]{https://www.kaggle.com/c/tiny-imagenet}
\footnotetext[4]{https://github.com/wenzhu23333/Federated-Learning}
\footnotetext[5]{http://kdd.ics.uci.edu/databases/kddcup99}
\footnotetext[6]{http://ufldl.stanford.edu/housenumbers}
\footnotetext[7]{https://github.com/xmouyang/FL-Datasets-for-HAR}
\footnotetext[8]{https://www.v7labs.com/open-datasets/office-caltech-10}
\footnotetext[9]{https://physionet.org/content/mimiciii-demo/1.4}
\footnotetext[10]{https://ai.bu.edu/M3SDA}
\footnotetext[11]{https://www.s.uci.edu/dataset/227/nomao}

\subsubsection{Heterogeneous federated transfer learning}
HEFTL mainly refers to the problem of inconsistency in feature or label space between participants in FL. To the specific, similar to HOFTL, it is assumed that there are two participants $i$ and $j$, whose private data constitute the source domain $\emph{D}_{i}$ and the target domain $\emph{D}_{j}$, respectively. HEFTL demonstrates the differences in either their domain ($\emph{D}_{i} \neq \emph{D}_{j}$) or task ($\mathcal{T}_{i} \neq \mathcal{T}_{j}$), which caused by their various feature spaces ($\mathcal{X}_{i} \neq \mathcal{X}_{j}$) or/and label space ($\mathcal{Y}_{i} \neq \mathcal{Y}_{j}$). Based on this, HEFTL has three scenarios:
\paragraph{Feature space heterogeneity: $\mathcal{X}_{i} \neq \mathcal{X}_{j}$}
\paragraph{Label space heterogeneity: $\mathcal{Y}_{i} \neq \mathcal{Y}_{j}$}
\paragraph{Feature and label space heterogeneity: $\mathcal{X}_{i} \neq \mathcal{X}_{j}$ and $\mathcal{Y}_{i} \neq \mathcal{Y}_{j}$}
Unlike HOFTL, where all participants train models with the same data structure, HEFTL allows for collaboration between datasets that are not identically structured. Thus, vertical federated learning is a prime case for HEFTL, since the local private data among participants in VFL may contain different sets of attributes or dimensions. Next, we give a detailed description of the settings as mentioned above.

\paragraph{Feature space heterogeneity}
Feature space heterogeneity refers to the situation that the feature space $\mathcal{X}$ of different participants is inconsistent, while the label space $\mathcal{Y}$ is consistent, particularly when different datasets involved in the training process have different sets of features. For example, in FL, two retailers are trying to identify fake reviews by local model prediction. Retailer $A$ has a feature space that includes review length, the number of purchases, and purchase history, whereas retailer $B$ utilizes review timing, user location, and account age as feature space. They all annotated their reviews with binary labels as ``true'' (0) or ``fake'' (1). Although the reviews obtained by different retailers have inconsistent feature space, these retailers still aim to leverage FL to enhance the predictive performance of their respective local models within a consistent label space.

\paragraph{Label space heterogeneity}
Label space heterogeneity refers to the situation where different participants have consistent feature space $\mathcal{X}$ but inconsistent label space $\mathcal{Y}$, which is the exact opposite of feature space heterogeneity. For example, in FL, two international e-commerce platforms are aiming to improve their recommendation systems. Each platform operates in a different region and thus has different product categories that are relevant to their local markets. Platform $A$ serves the Asian market and uses categories like ``Apparel'', ``Gadgets'', ``Furniture'', and ``Anime Merchandise''. Platform $C$ is based in North America, and uses labels such as ``Clothing'', ``Tech'', ``Home Improvement'', and ``Sports Equipment''. All two platforms collect user data including browsing time, click-through rates, purchase history, and search queries, which make up their consistent feature space. However, the way they categorize their products (labels) varies due to regional differences in terminology and market demand, leading to an inconsistent label space.

\paragraph{Feature and label space heterogeneity}
Feature and label space heterogeneity, indicates the feature space $\mathcal{X}$ and label space $\mathcal{Y}$ are both inconsistent among different participants. For example, there are two different specialty health clinics using federated learning to predict if patients will need to return for more treatment. Each clinic has its own set of measurements and outcomes. Clinic $A$ focuses on heart health, measuring things like heartbeat patterns and blood tests, and is concerned with whether patients might come back with heart issues. Clinic $B$ is a general clinic in a remote area, tracking health indicators like blood pressure and weight, and wants to predict if patients will return for any follow-up care or need a specialist. Each clinic collects different health information (different feature spaces) and has different categories for what counts as a patient needing to return (different label spaces). Considering the health indicators may be helpful in predicting patient's heart issues in the future. Thus, they want to use FL to build better prediction models without sharing sensitive patient data.

In summary, heterogeneous FTL may occur when there is inconsistency in the participants' feature spaces or label spaces. The existing research primarily focuses on FTL with heterogeneous feature spaces where only the feature spaces are inconsistent. Other heterogeneous situations in FTL remain worthy of deeper investigation.

\begin{figure*}[t]
  \centering
  \includegraphics[scale=0.25]{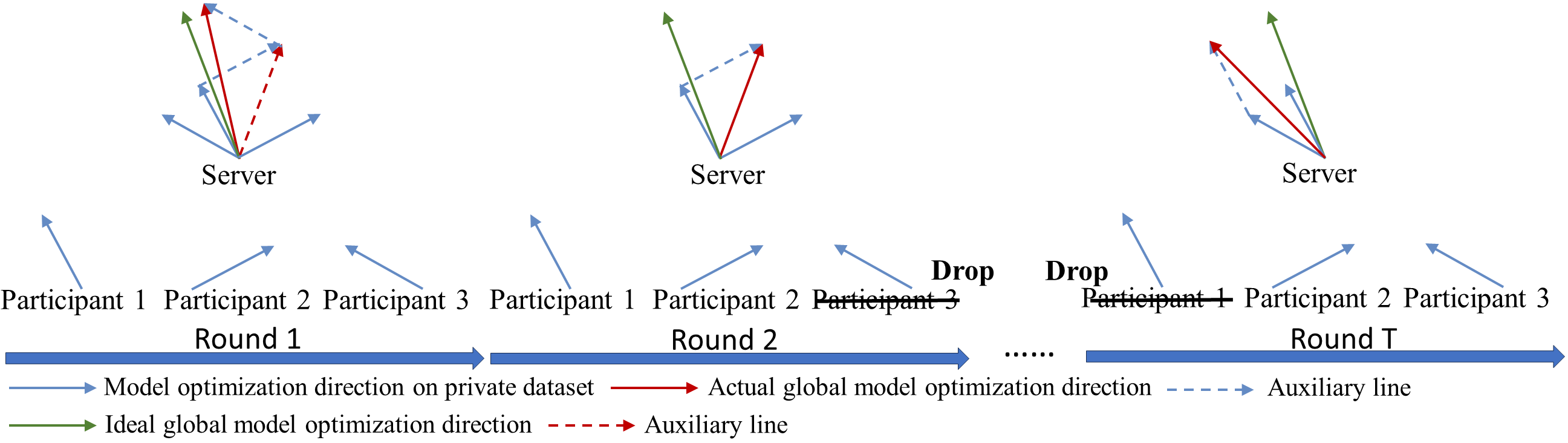}
  \caption{The detailed description of system heterogeneity in FTL. In each round of global communication, due to the resource heterogeneity among the participants, partial participants could not participate in the global aggregation in time, which results in the actual optimization direction of the aggregated model dynamically changing and deviating from the global optimal optimization direction.}
  \label{fig:system}
\end{figure*}
\subsection{Dynamic heterogeneous FTL}
DHFTL refers to the condition where the participant set that contributes to the FTL aggregation or the local raw data of partial participants in this set is dynamically changing at each round. We further provide detailed descriptions of the causes of dynamic heterogeneity.
\subsubsection{System heterogeneity}
Each participant's local device $F$ in FL could have different storage, computation, and communication abilities. Due to the varying storage or computational capabilities, some devices may not be able to complete the local training in time before aggregation. Meanwhile, the communication ability among participants is also influenced by network connections, and some devices may lose connection during a communication round because of connectivity or power issues \cite{li2020federated,chai2019towards,ye2023heterogeneous}. These aspects greatly amplify the straggler issue in the aggregation process \cite{schlegel2023codedpaddedfl}, forming a dynamically changing set of participants during FL iterations as shown in Figure \ref{fig:system}. Assuming that there is a global optimal direction $r^{g}_{e-1}$ for the global model aggregated by participants $u_{1},...,u_{n}$ in communication round $e$, and $m (0 \leq m \leq k, n \neq m)$ participants could send their local model to server in time due to device's limitation in communication round $e$, the global optimal direction $r^{g}_{e}$ aggregated by participants in round $e$ may have a significant difference with $r^{g}_{e-1}$ when there is data heterogeneity among participants, which is not conducive to the global model convergence. Therefore, how to transfer the knowledge among participants within a dynamically changing participant set is a key challenge for DHFTL. Dynamic heterogeneous FTL for participants $u_{1},...,u_{k}$ caused by system heterogeneity ($F_{i} \neq F_{j}, i,j \in (1,k)$) can be expressed as:
\begin{equation*}
    {D^{S}_{1},..., D^{S}_{n}}=\mathbb{S}_{e-1} \neq \mathbb{S}_{e}={D^{S}_{1},..., D^{S}_{m}},
\end{equation*}
where $n$ and $m$ represent the actual number of participants in communication round $e-1$ and $e$ of FTL, respectively. $\mathbb{S}_{e}$ indicates the set of actual participants in the communication round $e$.

\begin{figure*}[t]
  \centering
  \includegraphics[scale=0.25]{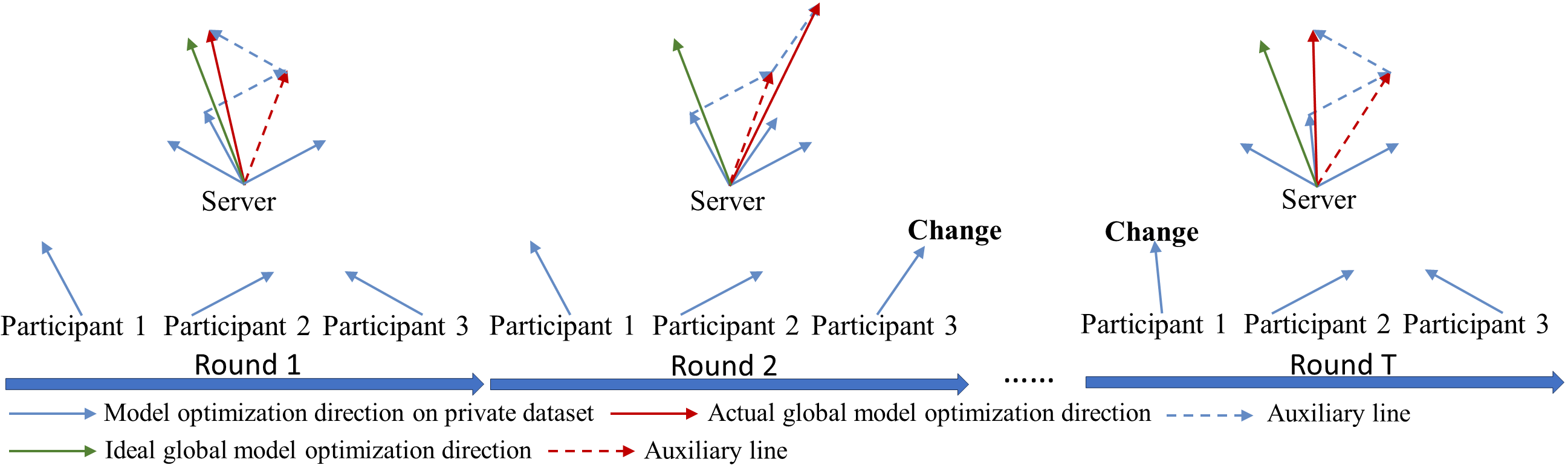}
  \caption{The detailed description of incremental data in FTL. In each round of global communication, due to the increase in the local user data or class, the local data distribution of participants may change, causing the actual optimization direction of the aggregated model to constantly vary and deviate from the global optimal optimization direction.}
  \label{fig:incremental}
\end{figure*}

\subsubsection{Incremental data}
Real-world FL applications are often dynamic, where local participants receive the new data, classes or tasks in an online manner \cite{you2022incremental,dong2022federated,chen2020fedhealth}, which proposes a key challenge is how to execute FTL from dynamically changing data distributions \cite{ditzler2012incremental,elwell2011incremental}. If only some participants are constantly adding data, or even if each participant synchronously adds new data, the newly added data could disrupt the original local data distribution, potentially exacerbating the differences between participant distributions as represented in Figure \ref{fig:incremental}. This requires the model to generalize well across both the old and new domains \cite{tan2020incremental,huang2023survey}. In addition, it's also possible that the feature space of the newly added data is inconsistent with the original feature space. Dynamic heterogeneous FTL for participants $u_{1},...,u_{k}$ caused by incremental data ($X^{i}_{e-1} \neq X^{i}_{e}$ or $\{\mathcal{X},\mathcal{Y}\}^{i}_{e-1} \neq \{\mathcal{X},\mathcal{Y}\}^{i}_{e}, i \in (1,k)$) in communication round $e$. Dynamic heterogeneous FTL in a participant can be written as:
\begin{equation*}
    P(X^{S}_{i,e-1}) \neq P(X^{S}_{i,e}) \: or \: \{\mathcal{X},\mathcal{Y}\}^{S}_{i,e-1} \neq \{\mathcal{X},\mathcal{Y}\}^{S}_{i,e}.
\end{equation*}
Another situation where dynamic heterogeneous FTL of multiple participants can be denoted as:
\begin{small}
\begin{equation*}
    P(X^{S}_{i,e}) \neq P(X^{S}_{j,e}) \: or \: \{\mathcal{X},\mathcal{Y}\}^{S}_{i,e} \neq \{\mathcal{X},\mathcal{Y}\}^{S}_{j,e}. (i,j \in (1,k)).
\end{equation*}
\end{small}
Nevertheless, we can only observe popularity in typical incremental learning approach \cite{ditzler2012incremental,elwell2011incremental,huang2023survey,tang2022using}, while these problems in incremental FTL receive relatively less attention. 

\subsubsection{Model adaptive FTL}
In practical scenarios, due to differences in training objectives, participants may employ different model architectures $\mathcal{M}$ for training \cite{alam2022fedrolex,li2021fedh2l}. Therefore, employing conventional aggregation methods for heterogeneous model's output representations or parameters, such as averaging participants' parameters in FedAvg \cite{mcmahan2017communication}, cannot effectively complete knowledge transfer between participants in FL. Additionally, even if the dimensions of the intermediate feature outputs are consistent across different models, the representational capacity of these features for local data could still vary, hindering performance improvement of the model in the target participant \cite{alam2022fedrolex,li2021fedh2l,tan2022fedproto,yu2020heterogeneous}. TL generally assumes that the model architectures in the source domain and the target domain are consistent, however, the inconsistency of models in FL poses a new challenge for the performance of target participants' tasks by aggregating process. This challenge, implementing effectively federated training in a model heterogeneous setting, is referred to as model adaptive FTL, which can be denoted as for participants $u_{1},...,u_{k}$: $\mathcal{M}_{i} \neq \mathcal{M}_{j}, i,j \in (1,k)$

\subsubsection{Semi-supervised and unsupervised FTL}
Real-world FL applications especially need to use unlabeled data more than others \cite{jin2020towards,NEURIPS2022_71c3451f}. On the one hand, in cross-device federated learning \cite{kairouz2021advances}, individual devices create a lot of unlabeled data, like photos, texts, and health record data from wearables. It's unrealistic to label all this data due to its large volume. On the other hand, cross-silo FL \cite{kairouz2021advances} involves businesses, where data labeling often needs special knowledge. This is common in finance and healthcare sectors. Labeling this data would be time-consuming and expensive. Thus, SSFTL and USFTL have caught the interest of some researchers \cite{jin2020towards,li2023class,liang2022rscfed}. Overall, SSFTL has two common scenarios: $\textcircled{1}$ only one participant has labeled data; $\textcircled{2}$ several participants each have a small amount of labeled data locally, where case $\textcircled{1}$ is often seen in VFL, where it's usually assumed that only one active party has data label information. USFTL in FTL refers to the scenario where all participants lack labeled information. The labeled data scarcity in FTL for participants $u_{1},...,u_{k}$ is denoted as: $\bigcup_{i}X_{i} = X^{L} \rightarrow 0, i \in (1,k)$.
\begin{figure*}[t]
  \centering
  \includegraphics[scale=0.26]{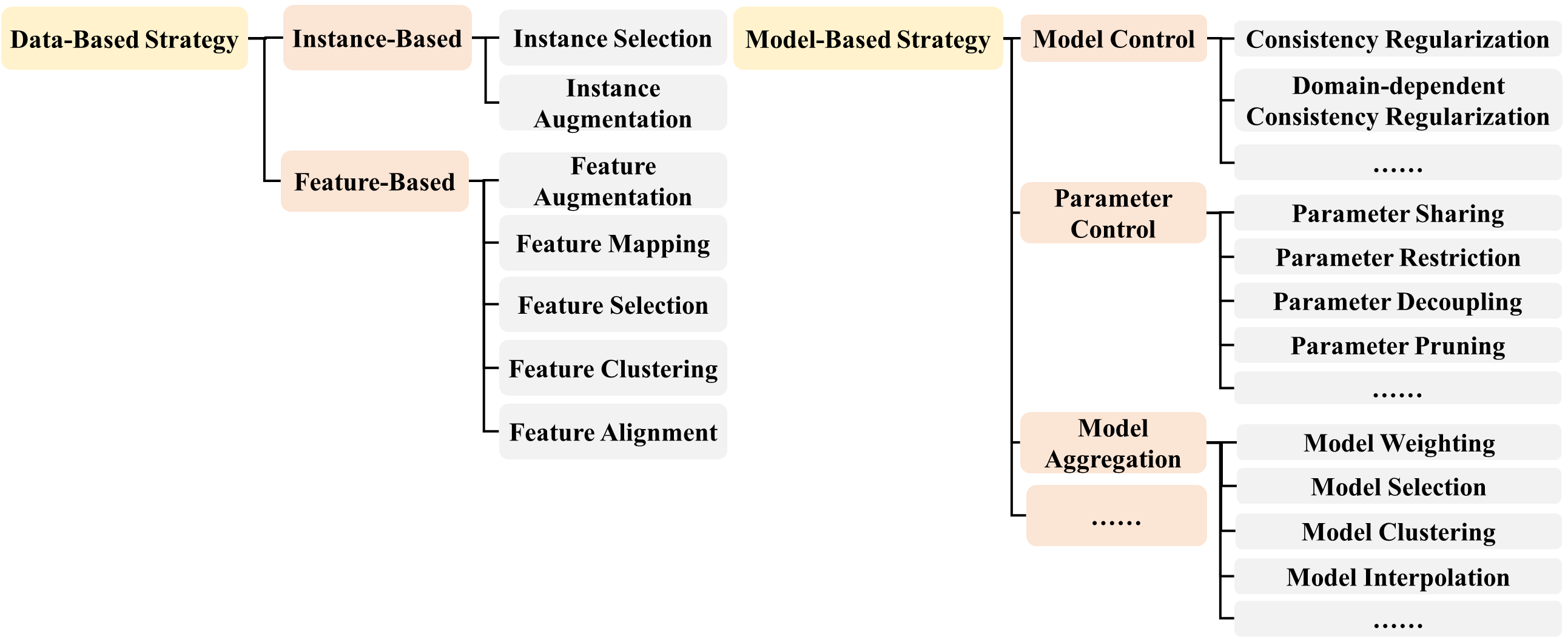}
  \caption{Data-based and model-based strategies of FTL}
  \label{fig:methodologies}
\end{figure*}

\section{Methodology}
We elaborate on the current research strategies for each FTL challenge mentioned in Section \ref{sec: FTL challenges}. As shown in Figure \ref{fig:methodologies}, it mainly includes two mainstreams: data-based and model-based strategies. Specifically, data-based strategies emphasize knowledge transfer by modulating and transforming participants' data for space adaptation, distribution adaptation, and data attribute preservation or adjustment \cite{zhuang2020comprehensive} without exposing any raw private data. The model-based strategies aim to improve the predictive accuracy of any given participant by the models from other participants in FTL. Table \ref{tab:study1}, \ref{tab:study2}, \ref{tab:study3} demonstrate related works on solving FTL challenges through these strategies. Note that currently there are very few FL works that specifically address the issues of label space heterogeneity or label \& feature space heterogeneity. Therefore, we will not discuss them in a separate subsection.
\subsection{Homogeneous federated transfer learning}
\label{sec:data methods}
Homogeneous FTL and heterogeneous FTL are two of the most studied challenges in FTL. We will first illustrate the strategies for addressing homogeneous FTL challenges from both data-based and model-based perspectives as shown in Figure \ref{fig:methodologies}. 

\subsubsection{Prior shift}
This subsection describes solutions to the prior shift challenge in HOFTL, which is one of the most common issue in FTL.
\label{subsubsec:ps}
\paragraph{Instance augmentation}
Instance augmentation in FTL aims to enhance data homogeneity of various participants through techniques like oversampling \cite{chawla2002smote,he2008adasyn,han2005borderline,wu2020fedhome,younis2022fly} and undersampling \cite{kubat1997addressing,yen2009cluster,tsai2019under}, which mainly occurs on the side of the participants. In detail, FedHome \cite{wu2020fedhome}, Astraea \cite{duan2020self}, and FEDMIX \cite{yoon2021fedmix} consider resolving prior probability bias at the local participant level. However, they ignore the effectiveness of global information in the local augmentation process. Therefore, some studies \cite{gong2022preserving,lin2020ensemble} suggest bridging the gap between participant and global distribution by creating a public dataset, but this also increases the risk of privacy leakage. To mitigate this issue, Faug \cite{jeong2018communication}, an FTL approach based on the generative adversarial network (GAN), is proposed to avoid privacy issues from multiple data transfers. Faug trains a GAN on minority class data at a central location, then sends it back to participants for data generation, helping build independent and identically distributed (IID) datasets. However, the construction of the generator increases extra computational and communication costs. the study \cite{hao2021towards} introduces a batch normalization (BN) based data augmentation approach, involving the following steps:
\begin{enumerate}
    \item BN layer parameterization: In the $\emph{t}$ round of global iteration, the $\emph{i}^{th}$ BN layer ($i \in {1,…,\emph{l}}$) of the global model $\mathcal{M}_{g}$ from the ${(\emph{i}-1)}^{th}$ round can be parameterized as a distribution of means $\mu_{i}$ and variances $\sigma_{i}$. For a given target category $\bar{y}(j) (1\leq j \leq \emph{z})$, where $\emph{z}$ represents the total number of possible categories, participants will sample from the Gaussian distribution to generate samples $\bar{x}(z)$ by forward propagate $\mathcal{M}(\bar{x}(j))$ ($1 \leq j \leq \emph{z}$), meanwhile the intermediate activation values $\emph{a}_{i}$ produced in sample process are applied to obtain the BN statistics $\bar{\mu}_{i},\bar{\sigma}_{i}$).
    \item Augmented Data Update: The computation of the loss function follows the formula: $\bar{x}(j)={argmin}_{\bar{x}}\sum^{l}_{i=1}{\Vert \bar{\mu}_{i}-\mu_{i} \Vert}^{2}_{2}+{\Vert \bar{\sigma}_{i}-\sigma_{i} \Vert}^{2}_{2} + \mathcal{L}_{\mathcal{H}}(\mathcal{M}$ $(\bar{x}(j)),\bar{y}(j))$, where $\mathcal{H}$ represents the cross-entropy loss. During the backpropagation process, the parameters of the model $\mathcal{M}$ are kept fixed, and only $\bar{x}(j)$ is updated to obtain augmented data that is closer to the real distribution.
\end{enumerate}
\paragraph{Instance selection}
Instance selection aims to select a subset of the available data that is most representative or informative for the training process. In distributed learning, numerous studies \cite{zhang2018crowdbuy,li2019todqa,katharopoulos2018not,alain2015variance,loshchilov2015online,schaul2015prioritized,wu2017sampling} suggest that constructing local training samples that are closer to the target distribution through instance selection methods can effectively enhance model performance. However, these methods are designed for traditional distributed training where training data is public. They are not suitable for FL, which uses private datasets from different owners. To solve this problem, the study \cite{Tuor2020DataSF} suggests using a benchmark model that is trained on a targeted small dataset before FL starts to evaluate the relevance of each participant's data, where only highly relevant data is used for local training. However, this method does not consider the influence of other participants on the benchmark model during FL training. Accordingly, the study \cite{shin2022fedbalancer} finds that calculating the sample loss during FL training can reflect the sample's homogeneity with the global data distribution. Among these, local samples with stronger homogeneity are more conducive to improving the global model's utility. As a result, this study proposes an FL framework named FedBalancer, which employs a selection strategy based on sample loss to filter local samples, aiming to build a local training set that is better aligned with the global sample distribution. However, FedBalancer increases computational costs because it requires calculating the loss value for every individual local sample. the study \cite{li2021sample} introduces a less computationally expensive method for selecting samples. This method uses the gradient upper bound norms of samples to assess their importance to global model performance. It calculates gradients from the loss of the last layer's pre-activation output, rather than calculating the gradient from the overall model parameters. This usually requires just one forward pass to accurately estimate a sample's importance. Specifically, the proposed algorithm includes the following two steps:
\begin{enumerate}
    \item Participant selection: using the private set intersection (PSI) protocol, each participant is informed about the target categories relevant to the target task. Participants with low relevance to the target task are filtered out. Reversely, if their total number of samples, which match the target categories, reach a certain threshold $d$, these qualified participants can participate in global aggregation. This prevents participants with large category distribution bias from interfering with the global model's convergence.
    \item Sample selection: during $t^{th}$ round, each participant $u$ of the selected participants $\bar{u}$ measures the importance $\lambda(x_{u,i},t)$ of samples $\{x_{u,i}\}_{i=1}^{n}$ related to the target task's categories, where $\lambda(x_{u,i},t)$ is defined as:
    \begin{equation*}
         \lambda(x_{u,i},t)=\sqrt{{\vert \sum_{u,l}\beta^{t,l}_{u,i}\nabla_{\alpha^{t,l}_{u,i}}f(x_{u,i};\theta_{t})} \vert}^{2},
    \end{equation*}
    where $\alpha^{t,l}_{u,i}, \beta^{t,l}_{u,i}$ are the input and output of the last layer ($l^{th}$) of sample $x_{u,i}$ in the $t^{th}$ iteration, respectively. $\sum_{t,l}\beta^{l}={\rm diag}({\delta'}^{l}(\beta_{1}),...,$ ${\delta'}^{l}(\beta_{r_{l}}))$. $\vert{\delta'}(\beta)\vert \neq \lambda$ and $f(x;\theta):=\sum^{k}_{u=1}\frac{n_{u}}{n}f_{u}(\theta)$. $\beta$ is the output matrix, $\delta$ is a trade-off parameter. The importance of a sample is indicated by the value of $\lambda(x_{u, i},t)$: higher values mean higher importance.
\end{enumerate}
Additionally, this selection strategy assumes that there are mislabeled samples locally, and these are often significantly more important than correctly labeled samples \cite{li2021sample}. Therefore, by filtering out outlier samples where $\lambda(x_{u, i},t)$ is significantly higher than most other samples, the above algorithm can effectively measure the true distribution of local categories, selecting samples closer to the global distribution for local model training.
\paragraph{Feature clustering}
Feature clustering seeks to find a more abstract representation of original features to group similar data distributions together \cite{zhuang2020comprehensive}. In the past, most clustering methods in FTL used model-related information for clustering \cite{long2023multi,briggs2020federated,sattler2020clustered,duan2020fedgroup,qayyum2022collaborative,mansour2020three,huang2019patient,ghosh2020efficient,ouyang2022clusterfl,li2021cbfl}. This model-related information includes things like model parameters \cite{long2023multi,briggs2020federated,ouyang2022clusterfl}, gradient information \cite{sattler2020clustered,duan2020fedgroup}, training loss \cite{ghosh2020efficient,mansour2020three,li2021cbfl}, and other external info \cite{qayyum2022collaborative,huang2019patient}. Except for these, clustering methods based on data-related information also can be applied to mitigate the prior shift issue. For example, Astraea \cite{duan2020self} traverses all unassigned participant data distributions by a greedy strategy, looking for a group of participants that can make the overall data distribution of each cluster as close to a uniform distribution as possible. 

\paragraph{Feature selection}
For example, Fed-FiS \cite{banerjee2021fed} generates local feature subsets on each participant by estimating the mutual information between features and between features and categories. Then, the server ranks each feature and uses classification tasks to obtain a global subset of features. Similarly, research \cite{cassara2022federated} utilizes a federated feature selection algorithm based on MI in the operation of autonomous vehicles (AV). This algorithm completes global iterative feature selection by locally executing an aggregation function based on Bayes' theory, which greatly reduces the computational cost. Moreover, 
Feature selection is a common idea to extract important features, which can obtain similar performance across different domains, and these important features can serve as a connection for knowledge transfer \cite{zhuang2020comprehensive}. In HOFTL, the local dataset of different participants may have similarities in feature space, and high dimensional features can delay the training time, leading to more energy consumption \cite{zhang2023federated}. In this case, removing irrelevant features and selecting useful overlapping features is crucial to address the distribution shift problem in FTL. Current FL researches \cite{zhang2023federated,hu2023federated,banerjee2021fed,cassara2022federated} have proposed a variety of solutions to the above problems, which mainly include three steps:
\begin{enumerate}
    \item Local filtering: filter the optimal subset of local features of each participant.
    \item Global filtering: aggregate local optimal feature subsets to obtain global feature set.
    \item Sharing: feed back the global feature set to the participants, allowing participants to focus on the features most relevant to the global representation.
\end{enumerate}
For example, FPSO-FS \cite{hu2023federated} is a federated feature selection algorithm based on particle swarm optimization (PSO), which proposes two global filtering strategies to determine the global optimal feature subset:
\begin{enumerate}
    \item Mean assembly strategy: for the private optimal feature subset of the $i$ participant, its average classification accuracy is obtained by the classification accuracy from all participants, i.e., ${\rm acc}_{ij},j=1,...,k$. Then, an optimal subset with the highest average classification accuracy is selected as the overall optimal feature subset, as follows: $\mathcal{X}^{*} = {\rm max}\{\mathcal{X}_{i}|\frac{1}{k}\sum^{k}_{j=1}{\rm acc}_{ij}(\mathcal{X}_{i},{\rm Dat}_{j}), i=1,...$ $,k\}$, where $\mathcal{X}_{i}$ is the private optimal feature subset from the $i^{th}$ $\emph{B}$ participant, ${\rm Dat}_{j}$ is the sample data held by the $j^{th}$ $\emph{B}$ participant, and $acc_{ij}(\mathcal{X}_{i}$, ${\rm Dat}_{j})$ is the classification accuracy of $\mathcal{X}_{i}$ evaluated by ${\rm Dat}_{j}$.
    \item Maximum and minimum assembly strategy: the first step is to identify the minimum classification accuracy for participant $i$'s optimal feature subset, using the classification accuracy ${\rm acc}_{ij},j=1,...,k$ obtained by all participants. Following this, the subset with the highest minimum classification accuracy among all private optimal subsets is selected as the overall optimal feature subset: $\mathcal{X}^{*} = {\rm max}\{\mathcal{X}_{i}|{\rm min}_{j=1,...,k} ({\rm acc}_{ij}(\mathcal{X},{\rm Dat}_{j})),i=1,...,k\}$.
\end{enumerate}
\paragraph{Consistency regularization}
The FTL atrategies also can be explained from a model perspective. Figure \ref{fig:methodologies} shows the corresponding strategies. Among them, consistency regularization \cite{zhuang2020comprehensive} refers to the addition of regularization terms to the objective function of local (or global) model optimization, which aims to improve the model robustness of participants, facilitating the transfer of knowledge from the source model to the target model during the training process. In traditional transfer learning, domain adaptation machine \cite{duan2009domain,duan2012domain} and consensus regularization framework \cite{zhuang2009cross,luo2008transfer} are widely used for knowledge transfer in multi-source domains \cite{zhuang2020comprehensive}, which are applicable to FL scenarios with two or more participants. The objective function is represented as:
\begin{equation*}
    {\rm min}_{f^{\emph{T}}}\mathcal{L}^{T,L}(f^{T}) + \delta_{1}\Omega^{D}(f^{T}) + \delta_{2}\Omega(f^{T}),
\end{equation*}
where the first term, as a loss function, is used to minimize the classification error of labeled target domain instances, the second term represents different regularizers, and the third term is used to control the complexity of the final decision function $f^{T}$. In addition, according to the research\cite{zhuang2020comprehensive}, domain-dependent consistency regularization can be expressed as:
\begin{equation*}
    \begin{aligned}
    {\rm min}_{f^{T}}\sum_{j=1}^{n^{T,L}}{(f^{T}(x_{j}^{T,L})-y_{j}^{T,L})}^{2} + \delta_{2}\Omega(f^{T})\\ 
    + \delta_{1}\sum_{u=1}^{k^{S}}\lambda_{u}\sum^{n^{T,U}}_{i=1}
    {(f^{T}(x_{i}^{T,U})-f_{u}^{S}(x_{i}^{T,U}))}^{2},
    \end{aligned}
\end{equation*}
where $\lambda_{u}$ represents the weighting parameter that is determined by the relevance between the target domain and the $u^{th}$ source domain. For example, pFedMe \cite{t2020personalized} utilizes Moreau envelopes for regularizing participants' loss functions. This approach effectively separates the optimization of individualized models from the learning process of the overarching global model within a structured bi-level framework tailored for FTL. MOON \cite{li2021model} proposes a contrastive learning-based federated optimization algorithm that uses the distribution difference in intermediate outputs between global and local models. Each participant's local optimization goal, beyond the cross-entropy loss term $\mathcal{L}_{u}^{T,L}(f^{T})$, aims to minimize the distance between the local and global model representations (reducing weight divergence) and maximize the distance between the local model and its previous version (accelerating convergence). 
\paragraph{Parameter sharing}
The parameters of a model essentially reflect the knowledge that the model has learned. Therefore, in FL, participants can also transfer knowledge at the parameter level \cite{zhuang2020comprehensive} by parameter sharing, which avoids the privacy risks brought by direct transmission of local data \cite{mcmahan2017communication}. For instance, the source and target models share parameters, and the target models use their local data to fine-tune the final layers of the source model, thereby creating a new model \cite{arivazhagan2019federated,wang2019federated,fallah2020personalized}. Since parameter sharing is a common approach in FL and often forms the basis for other methods, this section will focus on explaining the basic method of locally fine-tuning the global model. 

Specifically, study \cite{wang2019federated} finds that fine-tuning global model parameters with local training sets significantly improves prediction accuracy, particularly for local models that differ greatly from global predictions. Unlike FedPer \cite{arivazhagan2019federated}, which fine-tunes the global model's top parameters using local data, Per-FedAvg \cite{fallah2020personalized} averages all participant model parameters and fine-tunes all global model parameters using the MAML meta-learning method. However, the majority of existing FL frameworks based on parameter sharing mainly focus on improving the global model's performance on each participant using the participants' local data, overlooking the enhancement of a global model's generalization performance from the server's perspective. \cite{wang2023fedftha} proposes an FL framework based on a fine-tuning and head model aggregation method, called FedFTHA, which includes FedFT and FedHA. From the participant's perspective, FedFT focuses on improving the performance of the global model to the participant's local dataset by retaining and fine-tuning the local head model. From the server's perspective, FedHA works to reconstruct a global model that exhibits generalized performance, leveraging the participants' head model developed during FedFT. This approach enables both participants and the server engaged in FL to mutually benefit and realize a situation where all parties are advantaged.
\paragraph{Parameter restriction}
The knowledge learned from participants is kept as model parameters and is transferred by the server in CFL. Using the global model directly as the local model usually requires a strong correlation between the global and local data distributions. If there are large differences in data distribution among participants, using the global model directly and optimizing it with local data could lead to a significant decrease in the model's generalization ability. To address this, some studies \cite{li2020federatedoptimiztion,deng2020adaptive,t2020personalized,hanzely2020federated,dinh2020federated,chen2021bridging,hanzely2020lower,li2021ditto} in FTL restrict the similarity between the source and target models by parameter restriction \cite{zhuang2020comprehensive}. For example, FedProx \cite{li2020federatedoptimiztion} controls the differences between local and global model parameters by a proximal term, which aims to avoid the global model being significantly skewed by too many local updates and further affecting robust convergence. This approach keeps updates close to the initial model, helping to tackle the problem of prior distribution shift and covariate shift issues. However, this proximal term could not align local and global optimal points, and considering the potential loss of important parameter information when the global model is transferred locally. FedCL \cite{yao2020continual} introduces elastic weight consolidation (EWC) from continual learning \cite{kirkpatrick2017overcoming}. By using a server-side proxy dataset to estimate the importance of global model weights, local updates can be adjusted to prevent significant changes in vital parameters during local adaptive training:
\begin{equation*}
    {\rm min}_{f^{T}}\mathcal{L}_{u}^{T,L}(f^{T}) + \delta\sum_{i,j}\mathcal{R}_{ij} {\Vert \theta_{u,ij}^{T,L}-\theta_{g,ij}^{S,L} \Vert}^{2},
\end{equation*}
where $\mathcal{R}=\mathcal{R}_{ij}$ represents the importance matrix of the global model, derived using the server's proxy dataset. FedCL prevents divergence between global and local model weights and ensures better generalization and accuracy. Additionally, FedNova \cite{wang2020tackling} addresses distribution inconsistencies between participants by normalizing and scaling local updates, enhancing model convergence. SCAFFOLD \cite{karimireddy2020scaffold} focuses on reducing gradient variance, which first introduces a control variable $c_{u}$ for the direction of the participant model gradient, and then corrects local model updates based on the difference $(c-c_{u})$ between this control variable and a global variable $c (c=\{c_{u}\}_(u=1)^k)$, alleviating shift issue. Additionally, FedCSA \cite{ma2021fast} adjusts the weights of classifier parameters based on the distribution of each category on the participant. This adjustment enhances performance when dealing with class imbalance.
\paragraph{Parameter decoupling}
Research \cite{yu2020devil} indicates that the classifier of the model may exhibit significant accuracy decreasing when dealing with imbalanced prior probabilities. Therefore, many studies \cite{yu2020devil,kang2019decoupling,yosinski2014transferable,devlin2018bert} have suggested that sending partial local models for aggregation by decomposing models of participants into body (extractor) and head (classifier) parameters can improve accuracy in the target domain, which is called parameter decoupling. The body parameters can capture general data information and be maintained locally, enabling each participant to learn data characteristics for specific tasks, while head parameters learn specific features of the target domain for sharing with the server to improve the effectiveness of knowledge transfer. For instance, FedRep \cite{collins2021exploiting}, FedBABU \cite{oh2021fedbabu}, FedAlt \cite{pillutla2022federated}, FedPer \cite{arivazhagan2019federated} and SPATL \cite{yu2022spatl} perform global aggregation by sharing a homogeneous feature extractor. LG-FedAvg \cite{liang2020think}, CHFL \cite{liu2022completely}, and FedClassAvg \cite{jang2022fedclassavg} share a homogeneous classifier. Different from these, Fed-ROD \cite{chen2021bridging} shows great effectiveness by splitting the model head into general and personalized layers. Two predictors are trained using a shared body model to handle the competing objectives of generic and personalized federated learning. One predictor employs empirical risk minimization (ERM) for personalization, while the other uses balanced risk minimization (BRM) for general learning. Moreover, FedU \cite{zhuang2021collaborative} proposes a local sharing protocol based on a Siamese network. By aggregating only the online models from the source Siamese network to update the target model, it effectively enables knowledge transfer between participants. However, the study \cite{oh2021fedbabu} finds that existing parameter decoupling methods by updating the entire model during the training process, lead to a decrease in personalization performance. Therefore, it proposed an FL framework, called FedBABU. It updates only the body part of the model parameters during the federated training process, and the head is fine-tuned for personalization during the evaluation process. To more accurately determine the degree of impact each layer of the model has on the target domain, the study \cite{qu2023prevent} introduces a layered sharpness-aware Minimization (LWSAM) algorithm, which addresses the problem of poor participant performance due to the biased generic information shared by all participants. This method first calculates the distance between the global and local models at each layer, determining how much each layer is affected by the target domain. It accurately divides the model into head and body parts. Then it uses the sharpness-aware minimization (SAM) algorithm as a local optimizer. By adding more disturbances to the model body, the method adjusts the influence of the target distribution on the model from a global perspective.
\paragraph{Parameter pruning}
Due to varying data distributions among participants, directly applying an aggregated global model to a target domain often does not give optimal results. One popular solution \cite{yu2022spatl} is parameter pruning, which selects a subset of model parameters from the source domain to apply to the target domain. For example, FedMask \cite{li2021fedmask} uses a method called 'model binary masks' to selectively activate certain model parameters for training. This can happen after just one step of communication, without needing to fine-tune the model on local data, which reduces redundancy in communication and computation. In study \cite{yu2022spatl}, each participant uses a pre-trained reinforcement learning agent to choose parameters for combining data in a federated manner. They then use a global encoder and a local predictor to transfer knowledge from the combined model to individual models. Another study \cite{yang2022personalized} proposes a federated search method, which uses a lightweight search controller to find an accurate local sub-network for each participant. This method is good at extracting useful information and lowers the energy used for analysis and training. HeteroFL \cite{diao2020heterofl} proposes splitting the global model along its width while maintaining the full depth of the participants' DNN models, which aims to find more suitable local models. However, this approach could construct very thin and deep subnetworks, leading to a significant loss of basic features. To overcome this issue, the study \cite{ilhan2023scalefl} introduces a federated learning framework, named ScaleFL, which uses early exits to adaptively reduce DNN models' width and depth, finding models best suited for training with limited local resources.
\paragraph{Model weighting}
The knowledge transfer between participants can be accomplished by sharing local model-related information, such as model parameters, which involves aggregating them before local training, called model aggregation. However, different participants may have distinct optimal goals, simply averaging their model information with the same weight could result in the combined results not being the best solution \cite{mcmahan2017communication}. The global model in the server may also be overly influenced by a single participant's model, causing 'model drift' \cite{li2019convergence,wang2020tackling}. Thus, model weighting is applied to aggregate models according to their contributions. This prevents model performance degradation caused by directly averaging information from different actors into a domain. For example, the study \cite{zeng2021fedcav} finds local data with higher prediction errors has more contributions to improve the overall model performance, and then introduces an FL framework, called FedCav. FedCav measures the quality of local data using their prediction errors to decide the weights in the model aggregation process. Considering that the server doesn't know the local data distribution, FedFusion \cite{duan2023federated} uses a global representation of multiple virtual components with different parameters and weights to portray the data distribution of different participants. The server uses a variational autoencoder (VAE) to learn the best parameters and weights of the distribution components based on limited statistical information taken from the original model parameters. Additionally, the study \cite{liu2022deep} treats the blending of multiple models in FL as a graph-matching task, and then proposes an algorithm, called GAMF. It views channels and weights as nodes and edges of a graph, respectively. Then it uses a new hierarchical algorithm to increase the similarity of weights between channels, and proposes a cycle-consistent multi-graph matching method to merge various local source models in FL, enhancing the global model's generalization. Experiments show that GAMF can be used as a plug-in to boost the performance of existing FL systems. 
\paragraph{Model selection}
In reality, participants' local data may significantly differ from the optimal global distribution, and each participant's data characteristics can not be directly controlled, thus it is important to select participants related to the data or specific target labels for training, called model selection. 

For example, the study \cite{beilharz2021implicit} proposes a DFL algorithm based on directed acyclic graphs (DAG). In the DAG, each participant selects the model updates of other participants based on their data similarity, which has been demonstrated effectively in both prior and covariate shift scenarios. Astraea \cite{duan2020self}, a scheduler-based multi-participant rescheduling framework, re-schedules multiple participants through a scheduler, which follows that the data distribution of multiple participants is most similar to a uniform distribution. Dubhe \cite{zhang2021dubhe} is an FL algorithm based on repeated model selection, which allows the server to repeatedly select local models to participate in aggregation, and then send the encrypted distribution of the selected participants to the server to check the similarity between the global data distribution after aggregation and the consistency distribution. This process continuously adjusts the aggregation strategy and improves classification accuracy. Another study \cite{nagalapatti2021game} proposes a Shapley value-based federated averaging algorithm. It calculates the Shapley value of each participant to assess its relevance to the server's learning objective, estimating the participant's contribution in the next communication round of FL. This allows the server to select local models with higher contributions for training in each round of aggregation. In addition, some studies \cite{zhang2020personalized,li2022learning} require each participant to collect models from all other participants and use an additional local validation set to evaluate the similarity between participants. In contrast, the study \cite{liu2023feddwa} utilizes mathematical analysis methods instead of using empirical search from validation data sets to characterize the similarity between participants. Apart from data distribution, another study \cite{qi2023fedsampling} considers differences in local data volumes between participants. Note that uniform sampling could overlook participants with more data, reducing their contributions to the global model training and impacting the model's performance. To address this, the researchers propose FedSampling \cite{qi2023fedsampling}, a framework that uses a data uniform sampling strategy. When the participants' data distributions are highly imbalanced, participants randomly select others based on the ratio of the server's desired sample volume to the total available participant sample volume, further improving FL model performance.
\paragraph{Model clustering}
Some studies \cite{makhija2022architecture,mansour2020three} suggest that grouping similar participants for FL can address the model drift issue caused by data distributions' heterogeneity among participants, called model clustering. They determine participant similarity based on factors like model parameters \cite{huang2021personalized,mansour2020three,wang2020federated,briggs2020federated,tu2021feddl,yamada2020feature}, gradients \cite{duan2020fedgroup,sattler2020clustered}, training loss \cite{ghosh2020efficient,mansour2020three}, or other external information \cite{qayyum2022collaborative,huang2019patient}.

For example, FedCluster \cite{sattler2020clustered} uses cosine similarity of the model gradient to split participants into multiple clusters, maximizing similarity within clusters while minimizing it between clusters. To further enhance model adaptability in the target domain by utilizing the sub-model clustering method, the study \cite{zhu2022fednkd} designs a scale-based aggregation strategy, which scales parameters according to the pruning rate of the sub-models and aggregates overlapping parameters. It further introduces a server-assisted model adjustment mechanism to promote beneficial collaboration between device source models and suppress detrimental collaboration. This mechanism dynamically adjusts the sub-model structure of server devices based on a global view of device data distribution similarity. In addition, studies \cite{qayyum2022collaborative,huang2019patient} use exogenous information like the types of local devices participants use or patient drug features (drugs given within the initial 48 hours of ICU admission, comprising 1399 binary drug features) for clustering. However, these methods tend to overlook the 'cluster skew' issue caused by grouping, leading to the global model overfitting to a specific cluster's data distribution. ``Cluster skew'' refers not only to an imbalance in the category distribution among groups after clustering but also to an imbalance in the number of participants in each cluster. To address this issue, study \cite{nguyen2022feddrl} suggests a new FL aggregation method with deep reinforcement learning, called FedDRL, which can tap into the self-learning capability of the reinforcement learning agent, rather than setting explicit rules. Specifically, FedDRL utilizes a unique two-stage training process designed to augment the training data and reduce the training time of the deep reinforcement learning model. Moreover, the study \cite{hu2022spread} proposes a DFL framework based on hierarchical aggregation, named Spread. In this framework, the server acts as the FL coordinator, and edge devices are grouped into different clusters. Selected edge devices, as cluster leaders, responsible for model aggregation tasks. Spread monitors training quality and manages model aggregation congestion by adjusting both intra-cluster and inter-cluster aggregations. 
\paragraph{Model interpolation}
Different from model weighting methods that combine local models with varying weights, model interpolation aims to blend global and local model parameters proportionally to increase local prediction accuracy \cite{yoon2021federated,huang2021personalized,chen2021bridging}. For example, research \cite{hanzely2020federated} prevents the local model and global model from diverging excessively by setting an interpolation coefficient $\rho$ artificially. When $\rho$ is set to 0, the local model only performs local model learning; as $\rho$ increases, the local model gradually becomes similar to the global model, realizing mixed model learning; when $\rho$ is very large, all local models are forced to be similar, maximizing the transfer of knowledge from the global to the local model. Research \cite{marfoq2022personalized} interpolates a global model trained globally with a local k-nearest neighbors (KNN) model based on the shared representation, which is provided by the global model. The experiments show that it is also suitable for covariate issues except for prior shift issues. However, research \cite{mansour2020three} demonstrates that these methods, which rely on the separated training of the global model and local model to find the optimal interpolation coefficient $\rho^{*}$, may not always be the best. Therefore, it proposes a combined optimization strategy that improves both local and global models at the same time. Similarly, research \cite{chen2023elastic} proposes a model interpolation method based on elastic aggregation. They measure the sensitivity of each parameter by calculating the change in the overall prediction function output when each parameter changes, which reduces the update magnitude for more sensitive parameters, preventing the global model from excessively interfering with the local data distribution.

\subsubsection{Covariate shift}
\paragraph{Feature augmentation}
Some studies \cite{chen2023fraug,zhou2023fedfa} have proposed solving covariate shift issues from the perspective of feature augmentation. For example, the study \cite{chen2023fraug} proposes an FL paradigm based on a feature representation generator, called FRAug. It optimizes a common feature representation generator to help each participant generate synthetic feature representations locally, which are converted into participant-specific features by a locally optimized RTNet, which aims to make global and local feature distributions as similar as possible, increasing the training space for each participant. In addition, similar to the study \cite{hao2021towards} in data augmentation, the study \cite{zhou2023fedfa} further proposes FedFA to augment features from a statistical perspective. The statistics of augmented data should match as closely as possible with the statistics of the original training data. FedFA follows two preconditions:
\begin{enumerate}
    \item The data distribution of each participant can be characterized as the statistics of the latent feature distribution, i.e., mean and variance \cite{zhou2023fedfa}.
    \item The statistics of latent features can capture basic domain perceptual characteristics \cite{zhou2021domain,li2022feature,huang2017arbitrary}.
\end{enumerate}
Therefore, when dealing with data shift in FL, discrepancies in feature statistics across local participants are inconsistent, and they display uncertain changes compared to the actual distribution's statistics. Based on this, FedFA leverages a statistical probability augmentation algorithm based on a normal distribution to enhance the local feature statistics for each participant following as:
\begin{equation*}
    \bar{x} = \bar{\sigma}\frac{x_{u}-\mu_{u}}{\sigma_{u}}+\bar{\mu}_{u},
\end{equation*}
where the mean $\mu_{u}$ and variance $\sigma_{u}$ are the original statistics of the latent features, and $\bar{\mu}_{u}\sim\mathcal{N}(\mu_{u},\bar{\sum}_{\mu_{u}}^{2})$, $\bar{\sigma} \sim \mathcal{N}(\sigma,\bar{\sum}_{\sigma_{u}}^{2})$. $x_{u}$ is normalized by $x_{u}=\frac{(x_{u}-\mu_{u})}{\sigma_{u}}$, then expanded using the new statistical values $(\bar{\mu}_{u},\bar{\sigma}_{u})$. The variance $\bar{\sigma}_{u}$ dictates how much the latent features are augmented. The magnitude of this variance represents how much the latent feature distribution deviates statistically from the target distribution. By adjusting the variance $\bar{\sigma}_{u}$ appropriately, the skewness issue can be resolved either at the level of individual participants or across all participants. Moreover, it can be integrated as a plugin into any layer of any network to enhance features or solve any feature statistical bias, such as skewness in test time distribution. FedFA also exhibits excellent accuracy performance when addressing prior shift or quantity shift problems of homogeneous FTL.
\paragraph{Feature clustering}
PFA \cite{liu2021pfa} first clusters participants with similar data distribution by computing Euclidean Distance of local representations, and then orchestrates an FL procedure on a group basis to effectively achieve adaptation based on their clustering results. Additionally, the study \cite{huang2023rethinking} introduces a new FL framework, called FPL, which is based on prototype clustering. FPL uses the mean features of local data as prototypes, and clusters these prototypes at the server using a comparison learning method. This process brings similar prototypes closer and pushes different ones further apart. Moreover, to increase the stability of model training, FPL uses consistency regularization to minimize the distance between the representative prototypes and their unbiased counterparts. Compared to transferring model parameters in the FPL, the size of the prototypes is much smaller than that of the model parameters, which significantly reduces communication costs.
\paragraph{Consistency regularization}
From the model's perspective, directly adding model-level regularizers to the local objective function of the participants or server is a natural idea \cite{zhuang2020comprehensive}. In this way, the knowledge maintained in the model(s) of the participants (source model) can be transferred to the model of another participant (target model) during the training process. For fully supervised learning, each participant first uses their local labeled data to obtain the classification loss term. For example, FedBN \cite{li2021fedbn} keeps participant-specific batch normalization layers to normalize local data distribution. Its classification loss term can be written as:
\begin{equation*}
    \mathcal{L}^{T,L}(f^{T}) = \sum_{i=1}^{\bar{k}}\sum_{j=1}^{x_{i}}{(f^{T}(x_{j}^{i}),y_{j}^{i})}^{2}.
\end{equation*}
Except for the classification loss term, a consistency loss term can be introduced based on a cluster-aware mechanism, which uses the differences in both intermediate outputs and predictions between the global and local models to guide local model optimization \cite{wang2022fedkc}. It further groups the participants into clusters based on the feature clustering method by harnessing the similarity among lower-level features of each participant's model. Each cluster has its own global feature vector and average prediction value. By minimizing the L2 norm between each participant and the global feature and prediction value within its cluster, the method improves the robustness of local models under the covariate shift issue of homogeneous FTL. Additionally, PFL \cite{huang2023rethinking} introduces a consistency regularization term based on a global unbiased prototype. It suggests that the cluster prototype averaged by the server, as an unbiased prototype, can provide a relatively fair and stable optimization point. Calculating the square difference loss between the local feature vector and the unbiased prototype can address the issue of unstable prototype convergence. The regularizer in PFL can be expressed as:
\begin{equation*}
    \mathcal{L}_{{\rm regularizer}} = \sum_{j=1}^{v}{(x_{i,j}-\mathcal{U}_{j}^{k})}^{2},
\end{equation*}
where $i$ and $j$ index samples in dataset of participant $u$ and the dimensions of feature output, respectively. $v$ is the number of dimensions. $\mathcal{U}$ is the unbiased prototype.
\paragraph{Parameter decoupling}
Parameter decoupling is not only suitable for prior shift or quantity shift problems \cite{diao2020heterofl}, but also for solving covariate shift problems in homogeneous FTL. For example, the study \cite{li2021hermes} proposes a more flexible way of decoupling, designing an FL algorithm based on structured pruning, called Hermes. In this method, participants determine the sub-networks to participate in server aggregation through model pruning. To prevent information loss that could result from directly averaging local models, Hermes only averages overlapping sub-network parameters on the server, keeping the parameters of the remaining non-overlapping parts unchanged. The aggregated parts of the sub-networks are then sent back to the local devices for network updates, thereby improving the model's performance on local tasks.
\paragraph{Model weighting}
FedUReID \cite{zhuang2021joint} enhances the adaptability of the aggregated global model to each participant's local model by applying an exponential moving average (EMA) to update the global model for each participant, where the weight of the EMA represents the similarity between the global and local models. Additionally, FedDG \cite{zhang2023federateddomain} takes advantage of the domain flatness constraint, which serves as a substitute for the complex domain divergence constraint, to approximate the optimal aggregate weights. Moreover, FedDG uses a momentum mechanism to dynamically assign a weight to each isolated domain by tracking the domain generalization gap, improving its generalization capability. Past studies often simplify the blending of source models into a straightforward allocation problem, ignoring complex interactions between channels. Meanwhile, since model weights are shuffled during training, before merging, channels of each layer need to be aligned to maximize similarities in weights between multiple source models. This presents a quadratic assignment property problem, which is NP-hard problem. To tackle this problem, GAMF \cite{liu2022deep} propose to treat the channels and weights as nodes and edges of a graph to obtain the weights information. These weighting methods typically rely on model parameters or gradient differences to measure each participant's contribution to the target prediction. However, the transmission of this information involves potential privacy leakage risks. Thus, the study \cite{liu2023co} views the local model of each participant as black-box model, in which all data is stored locally and only the source model's input and output interfaces are accessible. Each participant's soft outputs are given a weight based on their inter-class variance. These weighted outputs are then used to create target pseudo-labels. Therefore, it proposes a federated adaptive learning framework called Co-MDA, called CO-MDA. CO-MDA changes the label noise learning section into a semi-supervised learning approach and proposes a Co2-Learning strategy. This strategy involves training two networks at the same time that filter each other's errors through epoch-level co-teaching \cite{han2018co}, and gradually co-guess the pseudo-labels with the outputs of both target networks to further reduce the impact of label noise. 

\paragraph{Model clustering}
FedDL \cite{tu2021feddl}, FedAMP \cite{huang2021personalized}, and HYPCLUSTER \cite{mansour2020three} use model-related information (such as model parameters, convolution layer channels, LSTM hidden states, and neurons in fully connected layers) to construct a shared global model based on model clustering method. However, these methods require several communication rounds to separate all inconsistent participants, potentially affecting computational and communication efficiency. Therefore, the study \cite{wang2020federated} proposes a method FedMA to achieve hierarchical clustering of participants with a single round of communication. This method uses the difference between the initial global model parameters and local model parameters to generate multiple sub-clusters. Then, by calculating the pairwise distances between participants within all sub-clusters, similar sub-clusters are iteratively merged until only a single cluster remains, containing all samples. Furthermore, similar to study \cite{long2023multi} in addressing prior shift issue, study \cite{xie2023federated} treats the multi-center participant clustering issue as an optimization problem, which can be effectively resolved using the expectation-maximization (EM) algorithm. In addition, different from traditional federated clustering methods, which associate each participant's data distribution with only one cluster distribution (known as hard clustering), the study \cite{ruan2022fedsoft} introduces a soft-clustering-based FL paradigm, called FedSoft. FedSoft allows each local dataset to follow a mixture of multiple cluster distributions, improving the training of high-quality local and cluster models.

\paragraph{Model selection}
Model selection, a classic transfer learning method, has seen widespread use in FTL, either on its own or in combination with other methods, and it is equally effective in addressing covariate shift issues of FTL. For example, CMFL \cite{luping2019cmfl} compares the local update of each participant with the global update during each iteration of learning by calculating the proportion of parameters in the local update that have opposite signs to those in the global update, which aims to assess the degree of alignment between the two sets of gradients. A higher proportion indicates a greater deviation from the direction of joint convergence, rendering the local update less relevant. CMFL thus selectively excludes such divergent local updates from being uploaded, effectively minimizing communication costs in FL while ensuring convergence can still be significantly achieved.

\subsubsection{Feature concept shift \& Label concept shift}
To mitigate feature concept shift challenges in FTL, study \cite{ghosh2020efficient} utilizes an iterative federated hierarchical clustering algorithm, called IFCA. Different from traditional methods, IFCA does not require centralized clustering algorithms. The server only plays a role in average model parameters, which substantially decreases the server's computational load. However, IFCA needs to run a federated stochastic gradient descent (SGD) algorithm in each round until it converges. This process could increase computational and communication efficiency in large-scale FL systems. 

Regarding the label concept shift issue, current studies address it from the perspectives of feature alignment \cite{liu2021feddg} or model clustering methods \cite{zhu2023confidence,zhu2023federated}. Feddg \cite{liu2021feddg} uses the amplitude spectrum in the frequency domain as data distribution information and exchanges it among participants. The goal is that each participant can fully utilize multi-source data distribution information to learn parameters with higher generalization, which proves equally effective under covariate shift. In addition, considering that Bayesian optimization is a powerful surrogate-assisted algorithm for solving label concept shift issues in FL and black-box optimization problems where the local model-related information is not visible to other participants for privacy. Some researchers have turned their attention to federated Bayesian optimization \cite{zhang2022personalized,dai2020federated,dai2021differentially,zhu2023federated}. However, these methods either have all participants work together on the same task, or make only one participant learn from others to tackle a specific task. However, in real life, the tasks of participants are often related to each other. the study \cite{zhu2023federated} introduces an efficient federated multi-task Bayesian optimization framework, called FMTBO, which dynamically aggregates multi-task models based on a dissimilarity matrix derived from predictive rankings. Additionally, FMTBO designs a federated ensemble acquisition function that effectively searches for the best solution by using predictions from both global and local hyperparameters, enhancing the generalization of the global model. Besides, pFEDVEM \cite{zhu2023confidence} introduces an FL framework based on Bayesian models and latent variables, and combines the model weighting strategy to mitigate label concept shift issues. In this setup, a hidden shared model identifies common patterns among local models, meanwhile, local models adapt to their specific environments using the information from the shared model, which determines the confidence levels of each participant. The confidence levels are then used to set the weights when combining local models. The extensive experiments have demonstrated that pFEDVEM robustly addresses three types of distribution shift issues including prior shift, covariate shift, and label concept shift, and obtains significant accuracy improvement compared to the baselines. 

\subsubsection{Quantity shift}
FEDMIX\cite{yoon2021fedmix} has proven that the instance enhancement method is equally effective for quantity shift problems. Moreover, studies \cite{donahue2021model,donahue2021optimality} view FL as a hedonic game, where each participant produces some cost (error) when joining in the FL process. There's a Nash equilibrium between minimizing individual errors and overall errors. For example, a school may aim to minimize its local error, while a region or city may aim to minimize the overall error. Study \cite{donahue2021model} proposes to find a relatively stable participant partition by accurately estimating the expected error of each participant, which may overlook the need to minimize the overall error. Additionally, the numbers of samples for participants in \cite{donahue2021model} are only assumed to be small or large, Different from it, study \cite{donahue2021optimality} not only more focuses on overall social well-being, but also is suitable for any number of participants with any various numbers of samples.
 
\subsection{Heterogeneous federated transfer learning}
This section will discuss existing works addressing the challenges of heterogeneous FTL, dynamic heterogeneous FTL, and model adaptive FTL from data-based and model-based perspectives as shown in Figure \ref{fig:methodologies}. However, it is worth noting that there is very little research on scenarios with heterogeneous label space and heterogeneous feature and label space, so we will not elaborate on it here.

\subsubsection{Feature space heterogeneity}
Heterogeneous feature spaces often occur in VFL, thus, we mainly focus on the VFL scenario for feature space heterogeneous FTL. To address this issue, researchers can utilize methods such as feature alignment \cite{gao2019privacy,liu2020secure,yang2020fedsteg} or feature concatenation \cite{wu2022practical} to construct new feature datasets for model training. Among them, feature alignment in VFL can be completed by constructing a novel feature subspace \cite{liu2020secure}, or filling in missing or incomplete features of each participant's feature spaces \cite{gao2019privacy}. These approaches enable knowledge transfer under a homogeneous feature space. Specifically, the study \cite{gao2019privacy} assumes an inconsistency in the feature spaces between two participants, the active participant $\emph{A}$ and the passive participant $\emph{B}$. Both of them map their features $\mathcal{X}_{com}^{B}$ using their respective mapping functions $\theta^{(A,B)}$ and $\theta^{B,A}$ to the same feature space, resulting in new feature representations $\mathcal{X}_{com}^{B} \cdot \theta^{(A,B)}$ and $\mathcal{X}_{com}^{B} \cdot \theta^{(B,A)}$. They optimize the mapping functions $\theta^{(A,B)}$ and $\theta^{(B,A)}$ by minimizing the similarity between the private features $\mathcal{X}_{pri}^{B}$ and $\mathcal{X}_{com}^{B} \cdot \theta^{(A,B)}$, as well as $\mathcal{X}_{pri}^{A}$ and $\mathcal{X}_{com}^{B} \cdot \theta^{(B,A)}$. Finally, participants $\emph{A}$ and $\emph{B}$, through secure bilateral computation, obtain the complete features $\mathcal{X}_{b}^{A}$ and $\mathcal{X}_{a}^{B}$ respectively. Based on this, the federated aggregation can be implemented under the aligned feature space. However, these methods rely on the existing feature space of participants, ignoring the relationships among these features. By combining feature clustering methods, the active participant can create new, more valuable feature space for knowledge transfer.
For example, study \cite{kang2022privacy} proposes a VFL paradigm based on feature space decomposition clustering, called PrADA. The specific steps include:
\begin{enumerate}
    \item Feature grouping: participant $\emph{C}$ applies domain expertise to divide raw features into $\emph{p}$ groups, each containing tightly related features. Moreover, participant $\emph{C}$ constructs $\emph{q}$ interactive feature among pairs of feature groups, resulting in a total of $\emph{h}$ feature groups (where $\emph{h} = \emph{p} + \emph{q}$).
    \item Pretraining stage: this stage involves collaborative efforts between source participant $\emph{B}$ and participant $\emph{C}$, to train a set of feature extractors ($f_{\mathcal{E}} = \{f_{\mathcal{E},\emph{i}}\}$ for $\emph{i}=1$ to $\emph{h}$) that are capable of learning features which are both invariant across domains and discriminative for labels.
    \item Fine-tuning stage: this process is executed in collaboration between active participant $\emph{A}$ and participant $\emph{C}$, with the goal of training participant $\emph{A}$'s target label predictor $\mathcal{C}$ by utilizing the pre-trained set of feature extractors ($f_{\mathcal{E}} = \{f_{\mathcal{E},\emph{i}}\}$, where $\emph{i}=1$ to $\emph{h}$).
\end{enumerate}
In addition, PrADA enhances privacy and security using a secure protocol based on partial homomorphic encryption. 

However, not all features of the participants are relevant to the task. Therefore, the novel feature space generated by aggregating the features of all parties needs to filter features that are not relevant to the task through feature selection. However, current feature selection techniques \cite{li2016deep,yamada2020feature} in distribution learning, often need numerous training iterations, particularly when dealing with high-dimensional data. Directly applying them in FTL to solve feature space heterogeneous issues produces significant computational and communication overhead, as each training round involves multiple encryptions, decryption operations, and intermediate parameter transfers. For example, study \cite{feng2022vertical} suggests using an embedded method to combine autoencoders with L2 constraints on feature weights for feature selection, and sets a threshold for post-training to determine the selected features for mitigating the problem of model parameter shrinkage \cite{louizos2017learning}. Different from previous VFL research scenarios where there were mostly two participants and binary classification tasks, study \cite{feng2020multi} proposes an FL feature selection scheme suitable for multi-participant multi-classification. In addition, previous studies that mainly focus on the relationship between features and labels \cite{jiang2022vf}, ignoring the relationship between features, to solve this problem, similar to research \cite{cassara2022federated,banerjee2021fed} using MI theory into federated feature selection in HFL, study \cite{fu2023feast} proposes a feature selection VFL framework based on conditional mutual information, called FEAST. FEAST integrates feature information into a single statistical variable for FL transmission, which not only accomplishes key feature selection and further reduces communication costs while ensuring privacy and security. In addition, study \cite{castiglia2023less} first proposes a theoretically verifiable feature selection method, formalizing the feature selection problem in the VFL environment, and providing a theoretical framework to prove that unimportant features have been removed.

\subsection{Dynamic heterogeneous FTL}
\subsubsection{System heterogeneity.} System heterogeneity among participants could lead to the emergence of stragglers in each iteration. To address this problem, instance selection can be leveraged by researchers to mitigate the computational burden of participants when they have heterogeneous local computational resources. However, it could lead to decreased model performance due to the reduced statistical utility of the training dataset. Study \cite{pilla2021optimal} obtains the optimal data selection scheme through an optimization function that includes lower and upper limits of resources, as well as arbitrary, non-decreasing cost functions per resource, meanwhile, it treats this problem as a scheduling problem of tasks assignment to resources, seeking to maximize the number of participants in each round of FL updates. In addition, FedBalancer \cite{shin2022fedbalancer} chooses samples for training by measuring their statistical utility, derived from the sample loss list based on the latest model. However, it is inefficient to wait for every participant to finish local training before proceeding with aggregation due to system heterogeneity. Thus, under a constant FL round deadline setting, instance selection could not immediately enhance the time-to-accuracy ratio. 
 
Furthermore, model-based strategies, such as consistency regularization \cite{li2020federatedoptimiztion}, model selection \cite{mcmahan2017communication}, model clustering \cite{li2022fedhisyn,xia2020multi,yang2021federatedreduction,yang2020age}, parameter decoupling\cite{chai2021fedat}, parameter pruning \cite{li2021fedmask} can also be applied as effective ways to address the straggler issue in FL. For example, Fedavg \cite{mcmahan2017communication} directly drops models of these stragglers which can not accomplish local training in time when the other participants have completed the same amount of training. Based on this, FedProx \cite{li2020federatedoptimiztion} allows varying local training epochs across participants, tailored to each device's system capabilities. Subsequently, it aggregates the non-convergent updates submitted by stragglers, rather than dropping these less responsive participants from this iteration. However, these frameworks may come at the cost of sacrificing accuracy due to the omission of partial information and waiting for all participants to complete a uniform number of training epochs tends to extend the convergence time of FL. FedAT \cite{chai2021fedat} blends synchronous and asynchronous updates by decoupling the model parameters at the layer level, which stratifies local models based on the time each participant needs to complete a round of training. During each training round, FedAT randomly selects some local models in each layer to calculate the loss gradient of local data, completing the synchronous update of models in that specific layer. Each layer, acting as a new training entity, then asynchronously updates the global model. The faster layers have shorter round-response delays, speeding up the convergence of the global model. The slower layers contribute to global training by asynchronously sending model updates to the server, which further improves the predictive performance of the model. Besides, \cite{li2021fedmask,yang2022personalized,ilhan2023scalefl} selectively use local models for transfer knowledge by parameter pruning methods under the limited computational resources. Study \cite{li2022fedhisyn} introduces a FL framework, called FedHiSyn, which uses a resource-based hierarchical clustering approach. This framework first categorizes all available devices according to their computing capabilities. Given that a ring topology is more suitable for models with uniform resources, after local training, the models are sent to the server. Then, within their respective categories, they exchange local model weight updates based on the ring topology structure to mitigate the lag effect caused by system heterogeneity. Considering that the main challenge of dynamic heterogeneous FTL is the appropriate scheduling of participants, essentially a local model selection issue at each iteration. Some researchers \cite{wang2020optimizing,chen2020joint,cho2020client,huang2020efficiency,nishio2019client,luping2019cmfl} assume the central party has 1-lookahead in source model selection strategies, which means that the dynamic input data beforehand is known. However, this can not be applied when dealing with unpredictable time series inputs. Thus, reinforcement learning has been increasingly used to design source model selection strategies \cite{wang2020optimizing,xia2020multi,deng2021auction,yu2022spatl,qu2022context}. Studies \cite{xia2020multi,yang2021federatedreduction} propose FL paradigms based on the multi-armed bandit (MAB). In situations where the available computing resources of participants are unknown, these approaches calculate the difference between the data distribution of multiple combined participants and a class-balanced data distribution, and then pick local models for aggregation and assign weights based on these differences. In another study, Study \cite{yang2020age} proposes a participant scheduling strategy by age of update (AoU) measurement. This strategy considers the age of the received parameters and the current channel quality at the same time, which improves efficiency and allows effective aggregation in federated joint learning. 

Except for the above-mentioned reinforcement learning methods in dynamic heterogeneous FTL, TiFL \cite{chai2020tifl}, a hierarchical federated learning framework, addresses system heterogeneity issues by grouping participants with similar training performance. Meanwhile, in each round of updates, TiFL adaptively selects local models for training within each group by simultaneously optimizing accuracy and training time. To accelerate model convergence under system heterogeneity, FedSAE \cite{li2021fedsae} suggests choosing participants with larger local training losses to take part in aggregation during each training round. FedSAE also designs a mechanism to predict each participant's maximum tolerable workload, aiming to dynamically adjust their local training rounds. Research \cite{yoon2022bitwidth} found that differences in bit-width among participant devices can impact the performance of the global model. Low-bit-width models are more compatible with hardware but could limit the model's generalization, leading to poor performance of high-bit-width models. To tackle this, the study \cite{yoon2022bitwidth} introduces ProWD, a FL framework that considers bit-width heterogeneity. This framework selects sparse sub-weights compatible with full-precision model weights from the low-bit-width models received by the server. These selected sub-weights then participate in central aggregation along with the full-precision model weights. Another study \cite{cox2022aergia} proposes a FL framework, called Aergia, which freezes the most computation-heavy parts of the model and trains the unfrozen parts. Moreover, the server chooses a more reasonable offloading solution based on the training speed reported by each participant and the similarity between their datasets. In this way, the training of the frozen parts can be offloaded to participants with ample resources or faster training speeds. PyramidFL \cite{li2022pyramidfl} is a fine-grained participant selection, which considers not only the distribution and system heterogeneity between the selected and non-selected participants but also within the selected participants themselves. Specifically, the server uses feedback from past training rounds to rank participants based on their importance, participants then use their rank to determine the number of iterations for data efficiency and the parameters to drop for system efficiency. Furthermore, the utility of each participant isn't static but varies across training rounds. If a participant is selected, its data utility will then decrease since these data have been seen by the model. Thus, reducing the likelihood of selection in subsequent training rounds allows participants who were not selected previously to have a higher probability of being chosen, which further improves the model performance under the dynamic heterogeneity and further enhances the fairness of participant selection.

\subsubsection{Incremental heterogeneity.} Existing FTL strategies mainly focus on model-based techniques, for example, GLFC \cite{dong2022federated} addresses the continuous emergence of new classes in federated online learning by consistency regularization method. It introduces a class-aware gradient compensation loss to ensure the consistency of the learning pace for new classes with the forgetting pace of old classes. It separately normalizes the gradients for new and old classes and reweights them for the local optimization goal, where the relationships between classes are obtained based on the best old classification model from the previous tasks. As the local data or tasks increase, where the tasks may be a new batch of data, the task $\mathcal{T}(t)_{i}$ learned by participant $u_{i}$ in round $t$ could be similar or related to the task $\mathcal{T}(t+1)_{j}$ learned by participant $u_{j}$ in round $t+1$. In this situation, the transmission of aggregated global information among participants can facilitate knowledge transfer across participants. Nonetheless, when a new joined task in participant $u_{j}$ is irrelevant to the tasks of participant $u_{i}$, it could affect the optimization direction of the local model, leading to a decrease in accuracy. To mitigate this issue, each participant selectively utilizes only the knowledge of the relevant tasks that have been trained on other participants during each iteration, while ignoring as much as possible the knowledge of irrelevant tasks that may interfere with local learning. Thus, parameter decomposition and model weighting methods have attracted the attention of researchers \cite{yoon2021federated}. For example, FedWeIT \cite{yoon2021federated} solves this problem by decomposing parameters into three different types for training: global parameters that capture the global and generic knowledge across all participants, local base parameters that capture generic knowledge for each participant, and task-adaptive parameters for each specific task per participant. Meanwhile, FedWeIT applies sparse masks to select parameters relevant to a given task, minimizing interference from irrelevant tasks of other participants and allocating attention to the server's aggregated parameters to selectively filter parameter information. However, all these methods lack a theoretical basis for ensuring convergence. FedL \cite{su2022online} uses dynamic adaptation to measure the extent to which online decision constraints are breached and calculates a maximum limit for this measure, which guarantees that the expected contribution of the chosen source model to the FL model's performance matches its actual contribution. 

\subsection{Model Adaptive FTL}
Model adaptive FTL is often caused by model heterogeneity, i.e., the local models of different participants may be inconsistent in architecture, which could cause incompatibility in the feature dimension and representational capacity among participants \cite{zhang2023towards}. It means that the average aggregation approach based on consistent features can not be used directly in FL. To mitigate this issue, data-based strategies are proposed in FTL. For example, the study \cite{liao2023draftfed} proposes to apply a feature mapping method to obtain consistent representation space and complete FL. Considering that even with different model structures, they possess some common knowledge for the same input, i.e., the feature extraction layers generate similar feature maps, research \cite{liao2023draftfed} uses ``model drafts'' to align local data distributions of participants. These outputs from specific layers or models are interpreted as blurred images of data and defined as model drafts. By minimizing the similarity difference between the local and global drafts, the data distribution difference between participants can be reduced. Except for feature mapping \cite{zhuang2020comprehensive,yue2022neural} using explicit features, some implicit features can be utilized to align the source and target domains, facilitating knowledge transfer within this aligned space \cite{zhuang2020comprehensive}, called feature alignment. Implicit features include subspace attributes \cite{liu2022feature}, spectral characteristics \cite{liu2021feddg}, prototype graphs \cite{tan2022fedproto}. For example, the study \cite{tan2022fedproto} uses prototypes to effectively transfer information under the model adaptive FTL by minimizing local and global prototype graphs within the same feature space, thereby capturing the semantic information of class structures. It avoids the possibility of data from different classes (across various participants) merging into a single class, or data from the same class being spread across multiple classes. Similarly, FedHeNN \cite{makhija2022architecture} is a FL framework based on instance-level representation. Each participant randomly selects part of local data and obtains instance-level representation to guide local training. By introducing a distance function based on centralized kernel alignment as a proximal term of the local loss function, it aligns local and global model representations, enabling federated learning across heterogeneous models. FedFoA \cite{liu2022feature} adds a linear calibration layer at the end of each local model to first calibrate the different feature dimensions among participants to the same dimension space. Participants use $QR$ decomposition to obtain feature subspace $R$ and feature correlation matrix $Q$. The central server minimizes the reconstruction of local features $\mathcal{X}$ and the product of global $R$ and initial vector $Q$ to get the optimal $Q^{*}$, guiding the update of the local sub-feature space. 

Inconsistent model architectures also make the simple average aggregation of model parameters ineffective. From the model-based perspective, studies \cite{diao2020heterofl,li2021hermes} implement FedAvg on top of local sub-networks by parameter decoupling. \cite{diao2020heterofl} assumes that the model architectures of participants can dynamically change with each iteration, and further proposes to leverage parameter decoupling to obtain at least one fixed sub-network for each type of heterogeneous situation and aggregate them into a single global model. Thus, smaller local models can gain more from global aggregation by performing less global aggregation on a subset of the parameters from larger local models. Similarly, study \cite{wang2023flexifed} combines parameter decoupling with model clustering method to group local models based on the similarity of their personalized sub-networks, maximizing the level of knowledge sharing between participants. Besides, research \cite{yang2023fedack} designed a Mapper at the local level to convert feature representations from different semantic spaces to the same feature space, and accomplish the knowledge transfer based on knowledge distillation (KD). They deploy a global generator at the server to extract global data distribution information and distill it into the local model of each participant. Then, local models are viewed as discriminators to reduce the difference between global and local data distributions in heterogeneous feature spaces. Since the feature representations synthesized by the global generator are usually more faithful and homogeneous to the global data distribution, they can achieve faster and better convergence. Additionally, local generators also can be used to enhance hard-to-judge sample data, improving model performance \cite{yang2023fedack}. In real-world scenarios, cross-institutional FL is often more content with the VFL scenario. However, traditional VFL can only benefit from samples shared among multiple parties, which severely limits its application. Therefore, research \cite{zhang2023towards} proposes a VFL framework based on representation distillation, called VFedTrans. This framework collaboratively models common features among multiple parties and extracts federated representations of shared samples, aiming to maximize data utility as much as possible through KD.

Knowledge distillation is also often used for dealing with model adaptive FTL induced by model heterogeneity as shown in Table \ref{tab:study3}. FedMD \cite{li2019fedmd} uses KD for federated learning in situations where different models are used. Instead of just combining model parameters, FedMD calculates class scores for each participant using a shared dataset. These scores are then sent to a server to calculate an average, which guides the training of local models. This method allows for knowledge sharing while keeping private data and model structures secure, and it works even when different local models are used. Contrary to the assumption that participants' local models are entirely different, research \cite{lin2020ensemble} assumes that local models of participants are not fully heterogeneous, and there are cases where some models share the same structure. Therefore, they propose an FL framework based on ensemble distillation, called FedDF. It creates several prototype models, which represent participants with identical model structures. In each round, FedAvg is performed among participants with the same prototype model to initialize a global model (student model), followed by cross-architecture learning through knowledge distillation. In this process, the parameters of the local model (teacher model) are tested on an unlabeled public dataset to generate predictions for training each student model on the server. Research \cite{huang2022few} adopts a federated communication strategy, denoted as FSFL, which is similar to FedMD, innovatively adding a latent embedding adaptive module to alleviate the impact of domain discrepancies between public and private datasets. However, these studies \cite{chang2019cronus,li2021fedh2l,gong2022federated,chen2020fedbe,li2019fedmd,lin2020ensemble,li2020practical,seo202216,sattler2020communication,wu2022communication,itahara2021distillation} strongly rely on the construction of public datasets, which undoubtedly compromises data privacy and is operationally challenging in practice. The impact of the quality of these prerequisites on the performance of federated learning is also unknown \cite{zhang2022fedzkt}. Therefore, research \cite{zhang2022fedzkt} proposes an FL framework based on zero-shot knowledge distillation, called FedZKT. FedZKT requires no prerequisites for local data, and its distillation tasks are assigned to the server to reduce the local workload. In addition, some studies \cite{zhu2021data,jeong2018communication,wu2021fedcg,zhang2022fine,zhang2023towards,yang2023fedack} introduce generators to avoid the need for public datasets, enabling the aggregation of local information in a data-free manner. For example, FedFTG \cite{zhang2022fine} uses the log-odds of each local model as a teacher to train a global generator and fine-tunes the global model using pseudo-data generated by the global generator. Due to the additional computational and communication costs imposed by the introduction of a generator, ScaleFL \cite{ilhan2023scalefl} proposes a self-distillation method based on exit predictions. This method treats self-distillation as an integral part of the local training process, requiring no extra overhead. ScaleFL enhances the knowledge flow among local sub-networks by minimizing the Kullback–Leibler divergence (KL divergence) between early exits (students) and final predictions (teachers). Furthermore, the study \cite{niu2023mckd} introduces a FL framework based on a data-free semantic collaborative distillation, called MCKD. This framework transfers soft predictions from local models to a server to learn representations that don't change across different domains. MCKD also introduces a knowledge filter to mitigate the potential amplification of irrelevant or malicious participants' influences on the target domain by traditional averaging aggregation. This knowledge filter generates consensus knowledge for unlabeled data and sets a threshold to drop models where local model predicted classes are inconsistent with consensus classes, further adapting the central model to target data. However, most of these methods construct ensemble knowledge by merely averaging the soft predictions of multiple local models, overlooking that local models have a differential understanding of distillation samples. Research \cite{wang2023dafkd} suggests that a model is more likely to make the correct predictions when the samples are included in the domain used for the model's training. Based on this, the study \cite{wang2023dafkd} proposes treating each participant's local data as a specific domain and designs a domain-aware federated distillation method named DaFKD. DaFKD can recognize the importance of each model to the distillation samples. For a given distillation sample, if the local model has a significant relevance factor with it, DaFKD assigns a higher weight to this local model.
 
\subsection{Semi-supervised and unsupervised FTL}
\label{sec:usftl}
The labeled data scarcity is a common scenario in both HFL and VFL scenarios. Data-based methods receive some attention in reality for SSFTL issues, some studies \cite{lin2021semifed,lubana2022orchestra} utilize these methods to obtain valuable augmented data by strategically optimizing predictions or similarity calculations on these data. For example, SemiFed \cite{lin2021semifed} uses both past local and global models to predict labels for local unlabeled data. When predictions have high confidence, the pseudo-labeled data is used to train as available data. Orchestra \cite{lubana2022orchestra} uses a feature clustering method to obtain more available data for training. It refines local data clusters through interactions with the server, which helps find effective samples in unlabeled data, further increases sample size, and reduces distribution heterogeneity when labeled data is scarce. Different from them without considering the high-class imbalance of unlabeled data, based on the instance selection method, CBAFed \cite{li2023class} uses the empirical distribution of all training data from the last round of global communication to design category-balanced adaptive thresholds. That is, if the model uses more data for training in one class, the threshold for labeling unlabeled data in this class will increase, and vice versa. It aims to shrink the gap between local and global distributions by selecting a local training set, and further influence category distribution, preventing a decline in global model performance due to prior probability bias.

Model-based strategies similarly show the effectiveness in addressing the SSFTL issue, study \cite{jeong2020federated} introduces a framework based on the consistency regularization method, called FedMatch. This framework splits local model parameters into two parts, one for updating with labeled data, and another for unlabeled data. For the parameters associated with labeled data, the loss function solely includes cross-entropy loss, and the loss function related to unlabeled data follows as:
\begin{equation*}
     {\rm min}_{f^{\emph{T,U}}}\mathcal{L}^{T,U}(f^{T,U}) + \delta_{L_{2}}{\Vert \theta_{L}^{*}-\theta_{U}\Vert}^{2}_{2} + \delta_{L_{1}}{\Vert \theta_{U}\Vert}_{1},
\end{equation*}
where $L_{1}-$ and $L_{2}-$ regularization on $\theta_{U}$ aims to make $\theta_{U}$ sparse, while not drifting far from the current optimal parameter $\theta_{L}^{*}$ trained with labeled data. The first term differs from the studies \cite{yang2021federated,lin2021semifed}, which not only uses the prediction results of unlabeled data and its augmented data to obtain cross-entropy loss but also considers pseudo-labels as real labels to obtain new category loss, making the model perform the same on original data and slightly perturbed data.

Contrary to previous studies that assume the presence of both labeled and unlabeled data locally, research \cite{jeong2020federated} describes a federated semi-supervised learning scenario where some participants have fully labeled data while others have only unlabeled data. To guide unlabeled participants' learning by building the interaction between the learning at labeled and unlabeled participants \cite{yang2021federated,lin2021semifed}, FedIRM \cite{liu2021fedirm} applies the intermediate output of the local model, which is trained by the labeled participants, to construct a category relationship matrix. By transmitting the relationship matrix of each labeled participant, it guides the unlabeled participants to learn their local relationship matrix. The formula is as follows:
\begin{equation*}
    {\rm min}_{f^{T,U}}\delta(w)(\mathcal{L}^{T,U}(f^{T,U}) + \mathcal{L}_{\rm{IRM}}),
\end{equation*}
\begin{equation*}
    \mathcal{L}_{\rm{IRM}} = \frac{1}{z}\sum_{j=1}^{z}(\mathcal{L}_{{\rm KL}}(\mathcal{R}^{L}_{j}||\mathcal{R}^{U}_{j})+\mathcal{L}_{{\rm KL}}(\mathcal{R}^{U}_{j}||\mathcal{R}^{L}_{j})),
\end{equation*}
where $\mathcal{R}_{j}$ denotes the relation vector of class $j$. Furthermore, researchers \cite{shen2020federated,li2021model} also apply domain-dependent consistency regularization to solve this issue. For example, \cite{shen2020federated} uses the KL divergence between local and global model predictions to control the update of the local objective function. Additionally, other model-based strategies, such as parameter decoupling \cite{jeong2020federated}, model weighting \cite{yang2021federated,liang2022rscfed,lin2022federated}, also attract some researchers' attention. FedMatch \cite{jeong2020federated} accomplishes the FTL under semi-supervised learning scenarios by training labeled and unlabeled data separately through model parameter decomposition. FedConsist \cite{yang2021federated} and RscFed \cite{liang2022rscfed} improve the performance of models in semi-supervised FTL by changing the weights of participants with labeled and unlabeled data. Since semi-supervised learning often deals with data that only has positive and unlabeled (PU) samples, one participant's negative class could be made up of multiple positive classes from other participants. However, traditional PU learning mostly focuses on binary problems with only one kind of negative sample. Thus, study \cite{lin2022federated} assumes that the available data has multiple types of positive and negative classes (MPMM-PU), and further proposes to redefine the expected risk of MPMM-PU, which aims to decide each participant's weight and examines the limits of the model's generalization.

In response to unsupervised FTL, a straightforward approach is to combine self-supervised methods with FL, such as \cite{jin2020towards,van2020towards}. However, this challenge is often accompanied by homogeneous or heterogeneous FTL issues. The feature-based strategies have obtained some attention to solve this problem, such as FedCA \cite{zhang2023unfederated} based on feature selection, FSHFL \cite{zhang2023federated} based on feature mapping, and FedFoA \cite{liu2022feature} based on feature augmentation. Moreover, studies also propose to alleviate the data drift issues between participants by model-based strategies, such as FedX \cite{han2022fedx} based on KD, and FedEMA \cite{zhuang2022divergence} based on model interpolation. FedCA \cite{zhang2023unfederated} shares features of local data and employs an auxiliary pubic dataset to minimize disparities in the representation space across participants. However, it ignores the inconsistency between the feature representation of local unlabeled data and global feature representation. To solve this problem, FSHFL \cite{zhang2023federated}, an FL framework based on an unsupervised federated feature selection approach, proposes the feature cleaning locally and global feature selection. The local feature cleaning utilizes an enhanced version of the one-class support vector machine (OCSVM) algorithm, called FAR-OCSVM, to identify features that lack sufficiently representative global features. The identification relies on local feature clustering, features within each cluster exhibit strong interrelationships, thus the clusters with more features contribute more significantly to the FTL process. Meanwhile, the server selects global features from the collected local feature sets and then returns these global features to participants, directing them to select local features that are closest to the global representation. Considering that the public dataset has potential privacy leakage risk \cite{zhang2023unfederated}, FedX \cite{han2022fedx} incorporates local knowledge distillation and global knowledge distillation into the FedAvg \cite{mcmahan2017communication} without any public data. Local knowledge distillation trains the network using the feature representations of local unlabeled datasets, and global knowledge distillation aims to mitigate data shift, which only relies on global model sharing. Besides, FedEMA \cite{zhuang2022divergence} utilizes self-supervised learning methods with predictors, including MoCo and BYOL, to update local encoders through the EMA of the global encoder. 

\begin{sidewaystable*}[]
\centering
\setlength{\tabcolsep}{1.4mm}
\renewcommand\arraystretch{0.8}
\begin{threeparttable}[b]
\caption{\centering FTL frameworks}
\begin{tabular}{|c|ccccccccccccccccccc}\hline
\bottomrule
                            & \multicolumn{13}{c|}{FTL challenges}                                                                                                                    & \multicolumn{4}{c|}{Architecture}                                                                  &  \multicolumn{1}{c|}{}                           \\\cline{2-18}
                                                  & \multicolumn{2}{c|}{DHFTL} &  \multicolumn{1}{c|}{}                &  \multicolumn{1}{c|}{}   & \multicolumn{1}{c|}{}                        & \multicolumn{5}{c|}{HOFTL}                       & \multicolumn{3}{c|}{HEFTL} & \multicolumn{1}{c|}{}               & \multicolumn{1}{c|}{}       & \multicolumn{1}{c|}{}                        & \multicolumn{1}{c|}{}                        &  \multicolumn{1}{c|}{}                           \\\cline{2-3} \cline{7-11} \cline{12-14}
\multicolumn{1}{|c|}{\multirow{-3}{*}{Reference}}  & \multicolumn{1}{c|}{System}   & \multicolumn{1}{c|}{Incremental}  & \multicolumn{1}{c|}{\multirow{-2}{*}{MAFTL}} & \multicolumn{1}{c|}{\multirow{-2}{*}{SSFTL}} & \multicolumn{1}{c|}{\multirow{-2}{*}{USFTL}} & \multicolumn{1}{c|}{PS} & \multicolumn{1}{c|}{CS} & \multicolumn{1}{c|}{FCS} & \multicolumn{1}{c|}{LCS} & \multicolumn{1}{c|}{QS} & \multicolumn{1}{c|}{FSH}    & \multicolumn{1}{c|}{LSH}    & \multicolumn{1}{c|}{FLSH}    & \multicolumn{1}{c|}{\multirow{-2}{*}{HFL}} & \multicolumn{1}{c|}{\multirow{-2}{*}{VFL}} & \multicolumn{1}{c|}{\multirow{-2}{*}{CFL}} &\multicolumn{1}{c|}{\multirow{-2}{*}{DFL}} & \multicolumn{1}{c|}{\multirow{-3}{*}{Strategy}} \\
\hline
\bottomrule
\end{tabular}
   \begin{tablenotes}
     \item[1] \scriptsize{Abbreviation: PS: prior shift; CS: covariate shift; FCS: feature concept shift; LCS: label concept shift; QS: quantity shift; FSH: feature space heterogeneity; LSH: label space heterogeneity; FLSH: feature and label space heterogeneity; HFL: horizontal FL; VFL: vertical federated learning; CFL: centralized FL; DFL: decentralized FL; IA: instance augmentation; IS: instance selection; FA: feature augmentation; FM: feature mapping; FS: feature selection; FC: feature clustering; FAI: feature alignment; CR: consistency regularization; DCR: domain-dependent consistency regularization; PS: parameter sharing; PR: parameter restriction; PD: parameter decoupling; PP: parameter pruning; MW: model weighting; MS: model selection; MC: model clustering; MI: model interpolation; KD: knowledge distillation.}
   \end{tablenotes}
   \label{tab:study1}
   \end{threeparttable}
\end{sidewaystable*}

\begin{sidewaystable*}[]
\centering
\setlength{\tabcolsep}{1.0mm}
\renewcommand\arraystretch{0.8}
\begin{threeparttable}[b]
\caption{\centering FTL frameworks}
\begin{tabular}{|c|ccccccccccccccccccc} \hline
\bottomrule
                            & \multicolumn{13}{c|}{FTL challenges}                                                                                                                    & \multicolumn{4}{c|}{Architecture}                                                                  &  \multicolumn{1}{c|}{}                           \\\cline{2-18}
                                                  & \multicolumn{2}{c|}{DHFTL} &  \multicolumn{1}{c|}{}                &  \multicolumn{1}{c|}{}   & \multicolumn{1}{c|}{}                        & \multicolumn{5}{c|}{HOFTL}                       & \multicolumn{3}{c|}{HEFTL} & \multicolumn{1}{c|}{}               & \multicolumn{1}{c|}{}       & \multicolumn{1}{c|}{}                        & \multicolumn{1}{c|}{}                        &  \multicolumn{1}{c|}{}                           \\\cline{2-3} \cline{7-11} \cline{12-14}
\multicolumn{1}{|c|}{\multirow{-3}{*}{Reference}}  & \multicolumn{1}{c|}{System}   & \multicolumn{1}{c|}{Incremental}  & \multicolumn{1}{c|}{\multirow{-2}{*}{MAFTL}} & \multicolumn{1}{c|}{\multirow{-2}{*}{SSFTL}} & \multicolumn{1}{c|}{\multirow{-2}{*}{USFTL}} & \multicolumn{1}{c|}{PS} & \multicolumn{1}{c|}{CS} & \multicolumn{1}{c|}{FCS} & \multicolumn{1}{c|}{LCS} & \multicolumn{1}{c|}{QS} & \multicolumn{1}{c|}{FSH}    & \multicolumn{1}{c|}{LSH}    & \multicolumn{1}{c|}{FLSH}    & \multicolumn{1}{c|}{\multirow{-2}{*}{HFL}} & \multicolumn{1}{c|}{\multirow{-2}{*}{VFL}} & \multicolumn{1}{c|}{\multirow{-2}{*}{CFL}} &\multicolumn{1}{c|}{\multirow{-2}{*}{DFL}} & \multicolumn{1}{c|}{\multirow{-3}{*}{Strategy}} \\
\hline
\bottomrule
\end{tabular}
   \begin{tablenotes}
     \item[1] \scriptsize{Abbreviation: PS: prior shift; CS: covariate shift; FCS: feature concept shift; LCS: label concept shift; QS: quantity shift; FSH: feature space heterogeneity; LSH: label space heterogeneity; FLSH: feature and label space heterogeneity; HFL: horizontal FL; VFL: vertical federated learning; CFL: centralized FL; DFL: decentralized FL; IA: instance augmentation; IS: instance selection; FA: feature augmentation; FM: feature mapping; FS: feature selection; FC: feature clustering; FAI: feature alignment; CR: consistency regularization; DCR: domain-dependent consistency regularization; PS: parameter sharing; PR: parameter restriction; PD: parameter decoupling; PP: parameter pruning; MW: model weighting; MS: model selection; MC: model clustering; MI: model interpolation; KD: knowledge distillation.}
   \end{tablenotes}
   \label{tab:study2}
   \end{threeparttable}
\end{sidewaystable*}

\begin{sidewaystable*}[]
\centering
\setlength{\tabcolsep}{1.0mm}
\renewcommand\arraystretch{0.8}
\begin{threeparttable}[b]
\caption{\centering FTL frameworks}
\begin{tabular}{|c|ccccccccccccccccccc}\hline
\bottomrule
                            & \multicolumn{13}{c|}{FTL challenges}                                                                                                                    & \multicolumn{4}{c|}{Architecture}                                                                  &  \multicolumn{1}{c|}{}                           \\\cline{2-18}
                                                  & \multicolumn{2}{c|}{DHFTL} &  \multicolumn{1}{c|}{}                &  \multicolumn{1}{c|}{}   & \multicolumn{1}{c|}{}                        & \multicolumn{5}{c|}{HOFTL}                       & \multicolumn{3}{c|}{HEFTL} & \multicolumn{1}{c|}{}               & \multicolumn{1}{c|}{}       & \multicolumn{1}{c|}{}                        & \multicolumn{1}{c|}{}                        &  \multicolumn{1}{c|}{}                           \\\cline{2-3} \cline{7-11} \cline{12-14}
\multicolumn{1}{|c|}{\multirow{-3}{*}{Reference}}  & \multicolumn{1}{c|}{System}   & \multicolumn{1}{c|}{Incremental}  & \multicolumn{1}{c|}{\multirow{-2}{*}{MAFTL}} & \multicolumn{1}{c|}{\multirow{-2}{*}{SSFTL}} & \multicolumn{1}{c|}{\multirow{-2}{*}{USFTL}} & \multicolumn{1}{c|}{PS} & \multicolumn{1}{c|}{CS} & \multicolumn{1}{c|}{FCS} & \multicolumn{1}{c|}{LCS} & \multicolumn{1}{c|}{QS} & \multicolumn{1}{c|}{FSH}    & \multicolumn{1}{c|}{LSH}    & \multicolumn{1}{c|}{FLSH}    & \multicolumn{1}{c|}{\multirow{-2}{*}{HFL}} & \multicolumn{1}{c|}{\multirow{-2}{*}{VFL}} & \multicolumn{1}{c|}{\multirow{-2}{*}{CFL}} &\multicolumn{1}{c|}{\multirow{-2}{*}{DFL}} & \multicolumn{1}{c|}{\multirow{-3}{*}{Strategy}} \\
\hline
\bottomrule
\end{tabular}
   \begin{tablenotes}
     \item[1] \scriptsize{Abbreviation: PS: prior shift; CS: covariate shift; FCS: feature concept shift; LCS: label concept shift; QS: quantity shift; FSH: feature space heterogeneity; LSH: label space heterogeneity; FLSH: feature and label space heterogeneity; HFL: horizontal FL; VFL: vertical federated learning; CFL: centralized FL; DFL: decentralized FL; IA: instance augmentation; IS: instance selection; FA: feature augmentation; FM: feature mapping; FS: feature selection; FC: feature clustering; FAI: feature alignment; CR: consistency regularization; DCR: domain-dependent consistency regularization; PS: parameter sharing; PR: parameter restriction; PD: parameter decoupling; PP: parameter pruning; MW: model weighting; MS: model selection; MC: model clustering; MI: model interpolation; KD: knowledge distillation.}
   \end{tablenotes}
   \label{tab:study3}
   \end{threeparttable}
\end{sidewaystable*}

\section{Application}
\label{sec:application}
In this section, we explore and outline the prevalent applications where FTL makes a significant impact. 
\subsection{Federated cross-domain recommendation}
Cross-domain recommendation (CDR) aims to reduce data sparsity by transferring knowledge from a data-rich source domain to a target domain. However, this process presents significant challenges related to data privacy and knowledge transferability \cite{liu2020exploiting,li2020ddtcdr}. Therefore, federated learning is introduced to CDR to improve the performance of the target domain model while providing privacy protection. Currently, classic recommendation algorithms have been widely applied to federated learning, such as federated collaborative filtering \cite{ammad2019federated,minto2021stronger}, federated matrix factorization \cite{chai2020secure,du2021federated,li2021federated}, and federated graph neural networks \cite{wu2021fedgnn}. However, these methods neglect the heterogeneity among them including data, resource, or model heterogeneity. Researchers have combined transfer learning techniques to accomplish tasks related to federated cross-domain recommendation. They utilize parameter sharing \cite{zhang2023dual}, parameter decoupling \cite{wu2021hierarchical,meihan2022fedcdr}, and model clustering \cite{luo2022personalized} to facilitate the process.
\subsection{Federated medical image classification}
Medical data involving patient information is sensitive and its use is strictly regulated, limiting the application of current artificial intelligence technologies in the medical field. Federated learning, which trains models on local devices without sharing raw data, could protect patients' data privacy security. However, data provided by different medical institutions, acting as source domains, often have heterogeneity in format, features, or distribution \cite{kaissis2020secure}. By integrating FL with transfer learning, we can leverage data from different healthcare institutions for model training, enhancing the model's performance while preserving data privacy. This approach presents a promising direction for the application of artificial intelligence in the medical field. Common methods include federated knowledge distillation \cite{sui2020feded,gong2022federated}, federated weighting aggregation \cite{silva2019federated,jin2021cross,xia2021auto,liu2023co,jiang2023fair}, federated consistency regularization \cite{acar2021federated,chen2022personalized}, federated model selection \cite{xu2022closing}, federated model interpolation \cite{jiang2023iop}, federated model decoupling methods \cite{wang2023feddp}, federated model clustering\cite{adnan2022federated}, and feature clustering methods \cite{wu2021federated}.

\subsection{Federated financial service}
Federated financial services include credit risk control \cite{li2023research,xu2023mses,lee2023federated,yang2019ffd,wang2024novel}, stock prediction \cite{pourroostaei2023federated,yan2022multi,shaheen2023reduction}, financial fraud transaction detection \cite{myalil2021robust,abadi2024starlit}, etc. Among them, credit risk control is a standard procedure for financial institutions, which estimates whether an individual or entity is able to make future required payments \cite{xu2023mses}. Stock prediction allows economists, governors, and investors to model the market, manage the resources and enhance stock profits \cite{pourroostaei2023federated}. Fraudulent transaction detection is another difficult problem for individual banks to curb company financial losses and maintain positive customer relationships \cite{myalil2021robust}. However, their predictive models all rely on large amounts of data to achieve training. Since the customers' data or fraudulent transaction data can not be shared among different institutions, the prediction model of each financial institution often suffers from the limited sample issue, which significantly hinders the performance improvement of the model. The emergence of federated learning not only breaks this “data silos” problem but also can combine instance augmentation \cite{yang2019ffd,shaheen2023reduction}, instance selection \cite{lee2023federated}, feature selection \cite{li2023research,xu2023mses,abadi2024starlit}, feature alignment \cite{liu2020secure,yang2020fedsteg}, parameter sharing \cite{wang2024novel}, model weighting \cite{pourroostaei2023federated}, model selection \cite{yan2022multi,myalil2021robust}, knowledge distillation \cite{wang2024novel}, and other technical approaches to achieve information exchange between institutions. 

\subsection{Federated traffic flow prediction}
Given that precise and up-to-date traffic flow data is of immense significance for traffic management, predicting traffic flow has emerged as a crucial element of intelligent transportation systems. However, current traffic flow prediction techniques, which rely on centralized machine learning, require the collection of raw data from mobile phones and cameras for model training, causing potential privacy issues. Researchers found that this issue also can be addressed using FTL, such as federated clustering aggregation \cite{liu2020privacy,zhang2021dual,zeng2021multi,zhang2021communication,qi2023fedagcn,phyu2023multi}, federated weighting aggregation \cite{xia2022short,zhang2022efficient,hu2023federatedride}, federated parameter sharing \cite{zhang2022efficient}, and federated parameter control methods \cite{phyu2023multi}.

\section{Conclusion and Future Work}
This survey provides a systematic summary of federated transfer learning, which emerges from the combination of transfer learning with federated learning, offering the corresponding definitions, challenges, and strategies. For convenience for researchers, we compile settings for the most common data heterogeneity scenarios, including both homogeneous and heterogeneous federated transfers, and summarize some significant FTL studies for various challenges. 

In the future, the utility performance of FL models in more complex scenarios deserves further exploration, including label concept shift, feature concept shift, label space heterogeneity, and feature \& label space heterogeneity. Due to differences or mutations in unknown or hidden relationships between input and output variables between participants, the concept shift becomes an important but less studied direction in FL. Meanwhile, label space heterogeneity and feature \& label space heterogeneity also deserve more in-depth study.

Furthermore, although federated transfer learning uses model selection, weight aggregation and other transfer learning methods to solve the problems of data heterogeneity, dynamic heterogeneity and labeled data scarcity in FL, new privacy concerns may have also emerged about these strategy preferences. After participants crack the weighting strategy or selection mechanism used by the server to aggregate information from all parties, they may use unfair means to make the training of the global model develop in a direction more beneficial to themselves. Thus, possible leakage of strategy preferences should be properly considered when designing FTL strategies in the future.

Finally, the communication costs, communication efficiency, and computational costs caused by these FTL strategies could be paid more attention by FL researchers. Compared with traditional FL methods, FTL strategies, such as instance augmentation, instance selection, feature selection, model selection, and model clustering, etc., increase additional computational costs and introduce more shared information among participants. It may lead to higher communication costs and affecting communication efficiency. In summary, FTL still has great potential research value in utility, privacy, and communication issues.


\bibliographystyle{fcs}

\end{sloppypar}
\end{document}